\documentclass{article}
\PassOptionsToPackage{numbers,sort&compress}{natbib}
\usepackage{iftex}
\ifPDFTeX
  \usepackage[utf8]{inputenc}
\fi
\usepackage[T1]{fontenc}
\usepackage{genai4health_2026}
\usepackage{hyperref}
\usepackage{url}
\usepackage{booktabs}
\usepackage{amsfonts}
\usepackage{amsmath}
\usepackage{amssymb}
\usepackage{nicefrac}
\usepackage{microtype}
\usepackage{graphicx}
\usepackage{xcolor}
\usepackage{colortbl}
\usepackage{tabularx}
\usepackage{enumitem}
\usepackage{algorithm}
\usepackage{algorithmic}
\usepackage{bbm}
\usepackage{makecell}
\usepackage{subcaption}
\usepackage[nohints]{minitoc}        %
\usepackage{tikz}
\usetikzlibrary{arrows.meta,calc,positioning}
\usepackage{pgfplots}
\pgfplotsset{compat=1.18}
\IfFileExists{fontawesome5.sty}{%
  \usepackage{fontawesome5}
  \newcommand{\dataseticon}{\faDatabase}
  
}{%
  \newcommand{\dataseticon}{Data}
  
}
\usepackage{multirow}
\usepackage{pifont}
\usepackage{xspace}
\hypersetup{hidelinks}

\newcommand{\xmark}{\ding{55}}
\newcommand{\cmark}{\ding{51}}
\newcommand{\pmark}{$\sim$}

\newcommand{\longcounsel}{\mbox{\textsc{LongCounsel-8}}\xspace}
\newcommand{\counselarc}{\textsc{CounselArc}\xspace}
\newcommand{\daic}{\textsc{Daic-Woz}\xspace}
\newcommand{\realcbt}{\textsc{RealCBT}\xspace}
\newcommand{\datasetqwenfull}{\textsc{LongCounsel-8-Qwen-3.5-35B-A3B}\xspace}
\newcommand{\datasetlunafull}{\textsc{LongCounsel-8-GPT-5.6-Luna}\xspace}
\newcommand{\datasetgptfull}{\textsc{LongCounsel-8-GPT-5.4-mini}\xspace}
\newcommand{\datasetqwen}{\textsc{LC8-Qwen}\xspace}
\newcommand{\datasetluna}{\textsc{LC8-Luna}\xspace}
\newcommand{\datasetgpt}{\textsc{LC8-GPT-5.4-mini}\xspace}
\newcommand\blfootnote[1]{%
  \begingroup
  \renewcommand\thefootnote{}\footnote{#1}%
  \addtocounter{footnote}{-1}%
  \endgroup
}
\newcommand{\datasetlink}{\href{https://huggingface.co/datasets/hiddensev/LongCounsel-8}{\nolinkurl{huggingface.co/datasets/hiddensev/LongCounsel-8}}}

\newcolumntype{L}{>{\raggedright\arraybackslash}X}
\newcolumntype{C}{>{\centering\arraybackslash}X}

\newcommand{\findingbox}[1]{%
  \par\vspace{4pt}
  \begingroup
  \setlength{\fboxsep}{4pt}%
  \setlength{\fboxrule}{0.4pt}%
  \noindent\fbox{%
    \parbox{\dimexpr\linewidth-2\fboxsep-2\fboxrule\relax}{%
      \normalfont \textbf{#1}%
    }%
  }%
  \par\endgroup
  \vspace{4pt}
}

\title{\longcounsel: A Benchmark Suite for Longitudinal Depression Tracking from Multi-Session Counseling Dialogues}

\author{%
  Jiayi Li,\;Bingsheng He,\;Zhaomin Wu \\
  Department of Computer Science\\
  National University of Singapore\\
  \texttt{li.jiayi@u.nus.edu, dcsheb@nus.edu.sg, zhaomin@nus.edu.sg}
}

\begin{document}

\doparttoc
\faketableofcontents

\maketitle

\begin{abstract}
Large language models (LLMs) are increasingly used to assess depression in counseling conversations, where psychology evidence suggests that tracking change across sessions matters more than any single score. However, existing dialogue benchmarks score each session on its own, and an accurate score for every session does not guarantee that a method follows the changes between sessions. We introduce \longcounsel, a benchmark for tracking depression across counseling sessions. Its systematic pipeline takes client profiles from real counseling cases, depression scores from real longitudinal data, and turn length and form from real counseling dialogues, yielding $7{,}749$ five-session trajectories from three LLMs. Every session is generated to express a preset score on the PHQ-8, a standard eight-symptom depression questionnaire. Validation shows that the labels are self-consistent and the dialogues close to real counseling. Following clinical research on treatment monitoring, we propose a trend-based evaluation protocol reflecting real-world tracking needs: changes between sessions, crossings of the screening cutoff, trajectory types, and symptom-level changes. Extensive experiments show all five existing methods tested fall short in tracking trends: none estimates changes better than a no-change baseline, at most 40\% of cutoff crossings are detected, and adding earlier sessions lowers change error but causes more missed improvements.

\blfootnote{\dataseticon\ Dataset: \datasetlink}
\end{abstract}

\section{Introduction}\label{sec:intro}

\begin{figure}[t]
  \centering
  \definecolor{exState}{HTML}{333333}
  \definecolor{exCur}{HTML}{2A78D6}
  \definecolor{exHist}{HTML}{EB6834}
  \definecolor{exBubble}{HTML}{F1F0ED}
  \begin{minipage}{0.6\linewidth}
  \centering
  \begin{tikzpicture}
    \begin{axis}[name=ex, scale only axis, width=0.74\linewidth, height=0.33\linewidth,
        xmin=0.55, xmax=5.45, ymin=2, ymax=16, clip=false,
        xtick={1,2,3,4,5}, xticklabels={S1,S2,S3,S4,S5}, xtick style={draw=none},
        ytick={5,10,15}, ylabel={PHQ-8 score},
        tick label style={font=\scriptsize}, label style={font=\scriptsize},
        axis x line*=bottom, axis y line*=left, axis line style={gray!70},
        legend style={draw=none, fill=none, font=\scriptsize, at={(0.5,1.06)}, anchor=south,
          inner sep=1pt, /tikz/every even column/.append style={column sep=8pt}},
        legend columns=2]
      \addplot[gray!60, dashed, line width=0.5pt, domain=0.55:5.45, samples=2, forget plot] {10};
      \addplot[exState, line width=1.1pt, mark=*, mark size=1.6pt, forget plot,
          mark options={fill=exState, draw=white, line width=0.3pt}]
        coordinates {(1,14) (2,12) (3,12) (4,14) (5,7)};
      \addplot[exCur, dashed, line width=1pt, mark=square*, mark size=1.5pt,
          mark options={solid, fill=exCur, draw=white, line width=0.3pt}]
        coordinates {(1,11.00) (2,9.00) (3,9.40) (4,11.80) (5,6.80)};
      \addlegendentry{EnsemBERT (Qwen3-Emb)}
      \addplot[exHist, densely dotted, line width=1.2pt, mark=triangle*, mark size=2pt,
          mark options={solid, fill=exHist, draw=white, line width=0.3pt}]
        coordinates {(1,9.60) (2,12.08) (3,11.68) (4,12.42) (5,11.51)};
      \addlegendentry{AIDA + history}
      \node[font=\tiny, text=exState, anchor=south, inner sep=1pt, yshift=2pt] at (axis cs:4,14) {controlled score};
      \node[font=\tiny, text=gray!55!black, anchor=west, inner sep=1pt] at (axis cs:5.45,10) {cutoff};
      \coordinate (t1) at (axis cs:1,2);
      \coordinate (t2) at (axis cs:2,2);
      \coordinate (t3) at (axis cs:3,2);
      \coordinate (t4) at (axis cs:4,2);
      \coordinate (t5) at (axis cs:5,2);
    \end{axis}
    \node[font=\tiny, anchor=east, inner sep=1pt] at ($(ex.south west)+(0,-0.6)$) {PHQ-8};
    \node[font=\tiny, anchor=east, inner sep=1pt] at ($(ex.south west)+(0,-0.88)$) {symptoms};
    \foreach \t/\total/\items in {t1/14/{2,2,3,3,1,1,1,1}, t2/12/{2,2,2,2,1,1,1,1}, t3/12/{2,2,2,2,1,1,1,1},
        t4/14/{2,2,3,3,1,1,1,1}, t5/7/{1,1,2,2,1,0,0,0}} {
      \node[font=\scriptsize\bfseries, inner sep=0pt] at ($(\t)+(0,-0.6)$) {\total};
      \foreach \s [count=\k from 0] in \items {
        \pgfmathsetmacro{\shade}{27*\s}
        \filldraw[fill=black!\shade, draw=gray!60, line width=0.2pt]
          ($(\t)+(-0.44+0.11*\k,-0.95)$) rectangle ++(0.1,0.14);
      }
    }
    \node[font=\tiny, anchor=west, inner sep=1pt] at ($(t5)+(0.5,-0.88)$) {(0--3)};
    \tikzset{quote/.style={fill=exBubble, rounded corners=3pt, inner sep=3pt, font=\scriptsize,
      align=left, text width={0.37\linewidth+0.22cm}}}
    \node[quote, anchor=north west] (q4) at ($(ex.south west)+(-0.9,-1.2)$)
      {\textbf{S4 client:} \emph{``i remember last night when i couldn't sleep and felt so hopeless.''}};
    \node[quote, anchor=north east] (q5) at ($(ex.south east)+(0.55,-1.2)$)
      {\textbf{S5 client:} \emph{``i sat alone at the cafeteria, eating fast to avoid anyone sitting next to me.''}};
    \node[font=\scriptsize\bfseries, anchor=north west, inner sep=0pt] (errtitle)
      at ($(ex.south west |- q5.south)+(-0.9,-0.22)$) {Mean absolute error on this trajectory};
    \begin{axis}[name=bars, at={($(errtitle.south west)+(1.65cm,-0.1cm)$)}, anchor=north west,
        scale only axis, width=0.74\linewidth-1.2cm, height=1.3cm,
        xbar=1.5pt, bar width=5.5pt, xmin=0, xmax=3.5, axis x line=none,
        ytick={0,1}, yticklabels={{Change MAE\\[-2pt]{\tiny over 4 changes}}, {Session MAE\\[-2pt]{\tiny over 5 sessions}}},
        enlarge y limits=0.45, axis y line*=left, y axis line style={gray!70}, ytick style={draw=none},
        yticklabel style={font=\scriptsize, align=right},
        nodes near coords, nodes near coords style={font=\tiny, text=black!85, anchor=west, xshift=1pt},
        nodes near coords={\pgfmathprintnumber[fixed, fixed zerofill, precision=2]{\pgfplotspointmeta}}]
      \addplot[fill=exHist, draw=none] coordinates {(3.06,0) (2.18,1)};
      \addplot[fill=exCur, draw=none] coordinates {(0.70,0) (2.20,1)};
    \end{axis}
  \end{tikzpicture}
  \end{minipage}
  \caption{An example of trajectory prediction on a \datasetqwen test trajectory, with symptom scores, client quotes, and prediction errors.}
  \label{fig:example}
\end{figure}

Large language models (LLMs) are increasingly used to assess mental health from language: they detect depression in social media posts, speech, and clinical interviews~\citep{lan2025depression,zhang2024llms,Chen2024DepressionDI} and complete psychiatric questionnaires from transcripts~\citep{rosenman2024llm,lee2026interpretable}. These assessments are typically evaluated one conversation at a time. Mental-health conversations, however, increasingly span many sessions, in online counseling and with AI chatbots, and risk accumulates over the course of an interaction~\citep{weilnhammer2026simvail}.

Psychology research suggests that tracking how a client's depression changes across sessions matters more than any single score. The change is how the effect of counseling is judged: improvement means a reliable and clinically meaningful change in a client's score~\citep{jacobson1991clinical,lowe2004monitoring}, and England's national psychological therapy service measures depression at every session to report how many clients improve~\citep{clark2018transparency}. Tracking the change also improves care: feeding repeated symptom scores back to clinicians improves treatment outcomes, whereas one-time screening does not~\citep{fortney2017tipping,guo2015measurement}. A method for multi-session counseling should therefore follow how a client's score on the PHQ-8, a standard depression questionnaire~\citep{kroenke2009phq8}, changes across sessions, which form a trajectory (Figure~\ref{fig:example}), rather than only score each session.

Methods that score each session accurately do not necessarily track the trajectory, and tracking therefore needs its own benchmark. This is because an estimated change is the difference of two session estimates and carries the errors of both. Letting these methods also read earlier sessions does not make them track changes better, because they then reduce the error of the estimated changes mainly by shrinking those changes, which flattens real changes. In Figure~\ref{fig:example}, AIDA + history, an existing method that also reads earlier sessions, scores each session slightly more accurately than EnsemBERT, yet it misses the client's improvement across the depression screening cutoff, and the error of its estimated changes is over four times higher. Evaluating tracking thus requires labels for every session of the same client and metrics on the changes between sessions. These metrics should reflect the questions that clinicians ask when they monitor treatment~\citep{lowe2004monitoring,fortney2017tipping}. For each client, a clinician asks how severe the depression is, whether it has improved or worsened, whether it has crossed the screening cutoff, and which symptoms have changed. A method can support such monitoring only if all of these questions are evaluated together on the same clients.

\begin{table*}[t]
  \centering
  \small
  \setlength{\tabcolsep}{4pt}%
  \caption{Data properties and evaluated dimensions (\cmark\ yes, \pmark\ partially, \xmark\ no; Section~\ref{sec:related} explains partial entries).}
  \label{tab:evaluation-gap}
  \begin{tabular}{@{}lc|ccccccc@{}}
    \toprule
    & & \multicolumn{7}{c}{\textbf{Evaluated dimensions}} \\
    \textbf{Resource} & \makecell{\textbf{Repeated}\\\textbf{labels}}
      & \makecell{\textbf{Current}\\\textbf{state}} & \makecell{\textbf{Change}\\\textbf{size}}
      & \makecell{\textbf{Change}\\\textbf{direction}} & \makecell{\textbf{Trajec-}\\\textbf{tory}}
      & \makecell{\textbf{Symptom}\\\textbf{change}} & \makecell{\textbf{History}\\\textbf{use}}
      & \makecell{\textbf{Dialogue-}\\\textbf{free}\\\textbf{baselines}} \\
    \midrule
    \multicolumn{9}{@{}l}{\textit{Longitudinal studies without dialogue}} \\
    MoodCapture & \cmark & \cmark & \xmark & \xmark & \xmark & \xmark & \xmark & \xmark \\
    PSYCHE-D & \cmark & \pmark & \xmark & \pmark & \xmark & \xmark & \xmark & \xmark \\
    \citet{Rou2026Contribution} & \cmark & \xmark & \xmark & \xmark & \cmark & \xmark & \pmark & \xmark \\
    \midrule
    \multicolumn{9}{@{}l}{\textit{Interview and counseling dialogue}} \\
    \daic & \xmark & \cmark & \xmark & \xmark & \xmark & \xmark & \xmark & \xmark \\
    PsychEval & \xmark & \xmark & \xmark & \xmark & \xmark & \xmark & \xmark & \xmark \\
    MusPsy & \xmark & \xmark & \xmark & \xmark & \xmark & \xmark & \xmark & \xmark \\
    TheraPhase & \pmark & \xmark & \xmark & \cmark & \xmark & \xmark & \xmark & \xmark \\
    \counselarc & \cmark & \cmark & \xmark & \xmark & \pmark & \xmark & \pmark & \pmark \\
    \midrule
    \makecell[l]{\longcounsel\\(this work)} & \cmark & \cmark & \cmark & \cmark & \cmark & \cmark & \cmark & \cmark \\
    \bottomrule
  \end{tabular}
\end{table*}

Existing resources rarely provide this combination (Table~\ref{tab:evaluation-gap}). The \daic interview corpus pairs one interview per participant with a PHQ-8 score and supports only current-state assessment~\citep{Gratch2014TheDA,burdisso2024daic}. Longitudinal health studies collect repeated questionnaire scores but no dialogue, and each evaluates a single aspect of change~\citep{Makhmutova2021PredictingCI,Rou2026Contribution}. Multi-session counseling corpora provide earlier dialogue but no session-level depression labels~\citep{na2026theraphase,pan2026psycheval}. \counselarc~\citep{wu2026emotrack} labels every session of its five-session counseling trajectories, but it has been evaluated only by session-level error.

We introduce \longcounsel, a benchmark for tracking depression across multi-session counseling dialogues (Figure~\ref{fig:pipeline}). Its systematic construction pipeline takes client profiles, depression scores, and dialogue form from real data, and three LLMs generate every session to express a preset PHQ-8 state. The resulting $7{,}749$ five-session trajectories have self-consistent labels and dialogues close to real counseling. Its trend-based evaluation protocol scores the same predictions not only on single sessions but also on the size and direction of changes, crossings of the screening cutoff, whole trajectories, and individual symptoms, against baselines that do not read the dialogue. Our contributions are as follows:
\begin{itemize}[leftmargin=*,itemsep=1pt,topsep=2pt]
    \item \textbf{A comprehensive benchmark.} We release \longcounsel, three validated datasets of five-session counseling trajectories with a PHQ-8 label for every session.
    \item \textbf{A trend-based evaluation protocol.} We propose seven metrics beyond session-level error that score changes between sessions, whole trajectories, and individual symptoms.
    \item \textbf{Key findings on existing methods.} Evaluating five representative methods yields five findings: (i) accurate session scores do not imply accurate tracking; (ii) most crossings of the screening cutoff are missed; (iii) fluctuating trajectories are often missed; (iv) changes in individual symptoms are not identified; and (v) adding earlier sessions lowers change error mainly by shrinking the estimated changes, which causes more improvements to be missed. These findings point future methods toward modeling change directly.
\end{itemize}

\section{Background}
\label{sec:background}

\paragraph{Measuring Depression with the PHQ-8.}
\longcounsel measures depression with the PHQ-8, a standard questionnaire that sums eight symptom scores into a total from 0 to 24~\citep{kroenke2009phq8}. It asks how often over the past two weeks a person has been bothered by each of eight symptoms: little interest or pleasure, low mood, sleep problems, tiredness, appetite changes, feeling bad about oneself, trouble concentrating, and slowed or restless movement. Each symptom is scored from 0 (not at all) to 3 (nearly every day). A total of 10 or more is the usual screening cutoff for at least moderate depression. The PHQ-8 is the PHQ-9 without its item on thoughts of self-harm; dropping that item turns a PHQ-9 response into a PHQ-8 response. We call a change of five or more points large, because it is the smallest change regarded as clinically meaningful in treatment monitoring~\citep{lowe2004monitoring}.

\paragraph{The \counselarc Construction Protocol.}
\label{subsec:chain}
\longcounsel builds on the construction protocol of \counselarc~\citep{wu2026emotrack} (Figure~\ref{fig:pipeline}), which fixes a PHQ-8 label for every session before the dialogue is generated; \counselarc itself was generated with Qwen3.5-35B-A3B. Each trajectory pairs a client profile derived from a real counseling case in PsychEval~\citep{pan2026psycheval} with five PHQ-9 totals from the longitudinal PSYCHE-D data~\citep{Makhmutova2021PredictingCI}; each total is matched to a real NHANES response with the same total~\citep{cdc_nhanes_dpq_l_2024}, whose first eight symptom scores form the session's controlled PHQ-8 state. Two LLM agents then role-play the session as counselor and client: the client shows its symptoms through pre-tested behavior cues, and turn lengths and forms are matched to the real counseling dialogues of \realcbt~\citep{Wang2025FeelTD}. After each session, the simulated client also fills out the PHQ-8 itself, which gives a second label, the self-report. \phantomsection\label{subsec:five-sessions}The five sessions follow PSYCHE-D measurements taken at months 0, 3, 6, 9, and 12, but the generation prompts specify only their order, not the time between them.

\section{Related Work}
\label{sec:related}

\paragraph{Resources for Depression Assessment.}
Most depression-assessment resources either collect repeated questionnaire scores without dialogue or provide counseling dialogue without a depression label for every session (Table~\ref{tab:evaluation-gap}). Longitudinal studies pair repeated questionnaire scores with smartphone or survey data~\citep{pfeiffer2015mobile,nepal2024moodcapture} and each evaluate one aspect: MoodCapture scores the current state, PSYCHE-D forecasts whether severity will increase over the next three months~\citep{Makhmutova2021PredictingCI}, and \citet{Rou2026Contribution} classify 12-week symptom trajectories and test the contribution of longitudinal features. LLM-generated counseling corpora provide dialogue at scale but mostly as single sessions without PHQ scores: CACTUS creates cognitive behavioral therapy (CBT) sessions from synthetic personas~\citep{lee2024cactus}, and MIRROR adds facial cues to therapy dialogues in which the client resists the counselor~\citep{kim2025mirror}. Multi-session corpora link sessions through treatment stages, evolving goals, memories, or case-derived client histories~\citep{na2026theraphase,wang2026muspsy,pan2026psycheval}, but they target tasks other than session-level depression assessment: TheraPhase predicts whether symptoms improve between the initial and completion stages of counseling against labels derived by an LLM, whereas PsychEval and MusPsy use symptom scales to evaluate counselor models rather than assessment methods. \counselarc~\citep{wu2026emotrack}, one of the three \longcounsel datasets, labels every session but reports only session-level error, broken down by trajectory type and history use, with a constant baseline for session MAE. Appendix~\ref{apdx:dataset-survey} gives an expanded survey.

\paragraph{Transcript-Based Depression Assessment.}
Transcript-based depression assessment methods mostly estimate severity from a single transcript and are evaluated one session at a time. They include supervised encoders and symptom-oriented predictors~\citep{lau2023automatic,milintsevich2023towards,ravenda2025transforming,agarwal2024analyzing}, as well as LLM-based methods that extract interpretable features, complete questionnaires, use in-context examples, or apply instruction tuning~\citep{lee2026interpretable,rosenman2024llm,Merzougui2025,Liu2025,Mental-LLM}.

\section{The \longcounsel Benchmark}
\label{sec:construction}

\paragraph{Design rationale.}
Real multi-session counseling transcripts with a depression label for every session are rarely publicly available (Table~\ref{tab:evaluation-gap}), and such data are difficult to share because of privacy concerns. \longcounsel therefore simulates clients whose PHQ-8 state is fixed before each session, so that the true size and direction of every change are known. To keep the simulated courses and conversations close to real counseling, the client profiles, score trajectories, and dialogue form are derived from real data (Section~\ref{subsec:chain}). Methods read only the transcripts and are compared with baselines that ignore the dialogue, so that any advantage over these baselines can be attributed to the dialogue.

\paragraph{Task.}
\longcounsel judges a method by how well it follows one client's depression across counseling sessions, not only by its score for each session, because the predictions for all five sessions of a trajectory are scored together (Figure~\ref{fig:example}). Each trajectory $i$ has five sessions with the same simulated client. Each session $t$ contains a dialogue of 20 counselor--client exchanges and two labels: the controlled PHQ-8 state, whose total $y_{i,t}$ is the evaluation target, and the client's self-report, which is the training label (Section~\ref{subsec:chain}). For each session, a method reads the transcript of session $t$ and, in history-using variants (marked +history), also the transcripts of sessions $1,\ldots,t-1$; it never sees an earlier label. It predicts a PHQ-8 total $\hat y_{i,t}$ and, optionally, eight symptom scores. Following~\citet{wu2026emotrack}, methods are trained and selected on the self-reports of the training and validation splits and evaluated against the controlled states of the test split.

\begin{figure*}[t]
  \centering
  \definecolor{pipelineNew}{HTML}{FDE4CF}
  \definecolor{pipelineNewEdge}{HTML}{D9541E}
  \tikzset{
    stage/.style={draw=gray!55, fill=gray!10, rounded corners=2pt, align=center,
      text width=1.6cm, minimum height=0.95cm, inner sep=2pt,
      font=\scriptsize, execute at begin node={\hyphenpenalty=10000\relax\exhyphenpenalty=10000\relax}},
    gate/.style={stage, fill=gray!22, text width=0.9cm},
    chip/.style={draw=pipelineNewEdge, fill=pipelineNew, line width=0.7pt, rounded corners=2pt,
      align=center, minimum width=2.05cm, minimum height=0.42cm, inner sep=1pt, font=\scriptsize},
    flow/.style={-{Latex[length=1.6mm,width=1.1mm]}, draw=gray!60!black, line width=0.6pt},
    newflow/.style={-{Latex[length=1.6mm,width=1.1mm]}, draw=pipelineNewEdge, line width=0.75pt},
    band/.style={rounded corners=3pt, inner sep=0pt}
  }
  \begin{tikzpicture}
    \fill[band, gray!4] (-0.2,1.3) rectangle (13.72,-1.0);

    \node[stage] (s1) at (0.87,-0.25) {\textbf{1. Client profile}\\[-1pt]{\tiny PsychEval cases}};
    \node[stage] (s2) at (2.95,-0.25) {\textbf{2. PHQ trajectory}\\[-1pt]{\tiny PSYCHE-D totals}};
    \node[stage] (s3) at (5.03,-0.25) {\textbf{3. Symptom vector}\\[-1pt]{\tiny NHANES responses}};
    \node[stage] (s4) at (7.11,-0.25) {\textbf{4. Behavior cues}\\[-1pt]{\tiny pruned by self-report}};
    \node[stage] (s5) at (9.19,-0.25) {\textbf{5. Dialogue}\\[-1pt]{\tiny two agents, \realcbt form}};
    \node[gate] (qc) at (10.92,-0.25) {Quality\\checks};
    \node[stage] (s6) at (12.65,-0.25) {\textbf{6. Self-report}\\[-1pt]{\tiny PHQ-8 after session}};

    \foreach \a/\b in {s1/s2, s2/s3, s3/s4, s4/s5, s5/qc, qc/s6} \draw[flow] (\a) -- (\b);

    \node[chip] (gq) at (6.84,0.85) {Qwen3.5-35B-A3B};
    \node[chip] (gl) at (9.19,0.85) {GPT-5.6 Luna};
    \node[chip] (gm) at (11.54,0.85) {GPT-5.4-mini};
    \draw[newflow] (gq.south) -- ++(0,-0.18) -| ([xshift=-3mm]s5.north);
    \draw[newflow] (gl.south) -- (s5.north);
    \draw[newflow] (gm.south) -- ++(0,-0.18) -| ([xshift=3mm]s5.north);

    \draw[newflow] (qc.south) -- ++(0,-0.30) -| node[pos=0.25, below, font=\tiny, text=pipelineNewEdge!85!black, inner sep=1pt] {fail: regenerate until pass} (s5.south);

    \fill[band, pipelineNew!35] (-0.2,-1.25) rectangle (13.72,-1.95);
    \draw[flow] (s6.south) -- (s6.south |- 0,-1.25);
    \node[font=\scriptsize, anchor=west, inner sep=0pt] at (-0.05,-1.6)
      {\textbf{\textcolor{pipelineNewEdge!85!black}{\longcounsel}}: three datasets of five-session trajectories, \datasetqwen\ (3{,}690), \datasetluna\ (3{,}690), \datasetgpt\ (369)};
  \end{tikzpicture}
  \caption{Construction pipeline of \longcounsel; orange marks additions to the protocol of~\citet{wu2026emotrack}.}
  \label{fig:pipeline}
\end{figure*}

\subsection{Data construction}
\label{subsec:construction}

\phantomsection\label{subsec:extensions}
\paragraph{Multiple generators.}
Three generator LLMs let us test whether dialogue properties and benchmark findings depend on the generator. We run the \counselarc protocol (Section~\ref{subsec:chain}) and its behavior cues with two further generators, GPT-5.6 Luna and GPT-5.4-mini. The benchmark therefore contains \datasetqwenfull~(\datasetqwen), which corresponds to \counselarc, together with \datasetlunafull~(\datasetluna) and \datasetgptfull~(\datasetgpt). The datasets differ in generator and decoding settings (Appendix~\ref{apdx:sampling}) and in size (Table~\ref{tab:corpus-statistics}).

\paragraph{Complete trajectories through regeneration.}
Every released trajectory has five complete sessions, each with both labels. The protocol checks every generated turn and self-report and repairs those that fail~\citep{wu2026emotrack}, and we additionally regenerate any session that remains invalid or is interrupted until it passes these checks. Appendix~\ref{apdx:regeneration} gives the attempt limits.

\begin{table*}[t]
  \centering
  \small
  \caption{Generators, statistics, and label fidelity of the three datasets.}
  \label{tab:corpus-statistics}
  \begin{tabular}{lccc}
    \toprule
    \textbf{Measure} & \textbf{\datasetqwen} & \textbf{\datasetluna}
      & \textbf{\datasetgpt} \\
    \midrule
    Generator & Qwen3.5-35B-A3B & GPT-5.6 Luna & GPT-5.4-mini \\
    \midrule
    Trajectories & 3{,}690 & 3{,}690 & 369 \\
    Client profiles & 369 & 369 & 369 \\
    Words/session & 826.1 & 988.5 & 927.5 \\
    Controlled total, mean (SD) & 6.13 (5.08) & 6.09 (5.12) & 6.24 (5.11) \\
    Improving / worsening & 491 / 281 & 478 / 311 & 65 / 27 \\
    Stable / fluctuating & 1{,}843 / 1{,}075 & 1{,}841 / 1{,}060 & 179 / 98 \\
    Train / validation / test & 2{,}583 / 369 / 738 & 2{,}583 / 369 / 738
      & 258 / 37 / 74 \\
    \midrule
    \multicolumn{4}{@{}l}{\textit{Label fidelity: self-report scored against the controlled state}} \\
    Session MAE $\downarrow$ & 0.521 & 0.294 & 0.541 \\
    Change MAE $\downarrow$ & 0.626 & 0.311 & 0.434 \\
    Trajectory-type agreement $\uparrow$ & 0.921 & 0.954 & 0.946 \\
    \bottomrule
  \end{tabular}
\end{table*}

\phantomsection\label{subsec:statistics}
\paragraph{Composition and splits.}
We split each dataset into training, validation, and test trajectories at a 7:1:2 ratio, and claims about new clients rest on profile-disjoint splits, in which no profile appears in two splits. The same profile can appear in more than one split of \datasetqwen and \datasetluna, because a profile can be paired with several independently sampled severity trajectories. \datasetgpt contains one trajectory per profile and is therefore profile-disjoint, and we additionally evaluate a 369-trajectory profile-disjoint split of \datasetluna.

\subsection{Evaluation protocol}
\label{subsec:protocol}

Our evaluation protocol considers three complementary perspectives: assessment at the session level, across sessions, and within each session. \textbf{At the session level}, we treat each session as a whole and assess the client's overall depression severity, measured by the PHQ-8 total. We evaluate both current-severity accuracy (\textbf{Session}) and whether earlier-session context improves this estimate (\textbf{Context}). \textbf{Across sessions}, we examine how the same client's severity evolves over time, assessing changes between adjacent sessions (\textbf{Transition}) and overall patterns along the full trajectory (\textbf{Trajectory}). \textbf{Within each session}, we look beyond the total score to the eight individual symptoms that compose it, since similar totals can reflect different symptom profiles. Symptom-level evaluation (\textbf{Symptom}) therefore tests whether methods identify changes in these individual symptoms that total scores can obscure. Together, these three perspectives give rise to five evaluation components. Each component is designed for a judgment clinicians make when they monitor treatment~\citep{lowe2004monitoring}. All metrics are computed from the same saved predictions, without additional training; MAE-based metrics are better when lower, and accuracy-based metrics when higher.

\begin{enumerate}[leftmargin=*,itemsep=1pt,topsep=2pt]
    \item \textbf{Session: current severity.} Monitoring starts from how severe the client is now. Session MAE, the mean absolute error (MAE) over all test sessions, measures this state.
    \item \textbf{Transition: change between sessions.} Monitoring must catch whether a client improves or worsens and whether the score crosses the screening cutoff. Such crossings matter because they change the client's clinical category. A drop below the cutoff is how routine outcome reporting defines recovery~\citep{clark2018transparency}, and a rise above it indicates probable depression~\citep{kroenke2009phq8}. Change MAE averages $|(\hat y_{i,t}-\hat y_{i,t-1})-(y_{i,t}-y_{i,t-1})|$ over the four adjacent transitions. Direction accuracy considers only large transitions, in which the controlled total changes by at least five points (Section~\ref{sec:background}), and counts a transition as correct when the predicted change has the same sign; a predicted change of zero counts as an error, and we report worsening and improvement separately. Threshold detection considers only transitions in which the controlled total crosses the cutoff of 10 and counts a transition as detected when the predictions cross in the same direction at the same transition, separately for upward and downward crossings.
    \item \textbf{Trajectory: overall pattern.} Some courses, such as a fluctuation that ends where it began, are visible only across all sessions. Type recovery applies the rule that labels controlled trajectories as improving, worsening, fluctuating, or stable (Appendix~\ref{apdx:trajectory-types}) to the five predicted totals and reports macro-F1 and per-type recall. Stratified MAE reports session MAE for each trajectory type, with comparisons at matched severity that separate the effect of trajectory type from that of severity.
    \item \textbf{Symptom: which symptoms changed.} The same change in the total can come from different symptoms. For methods that output symptom scores, symptom change MAE and symptom direction accuracy, the latter for symptom changes of at least one point, test whether a method identifies which symptoms changed.
    \item \textbf{Context: use of earlier sessions.} A method that reads earlier sessions should use them to estimate the client's state and changes better. We compare session MAE and change MAE as a method reads more earlier sessions (the history window); a method that uses history well lowers both. Because a method can also lower change MAE by shrinking its estimated changes, we read this comparison together with direction accuracy.
\end{enumerate}

\paragraph{Dialogue-free baselines.}
Every metric is compared with baselines that ignore the dialogue, and a method counts as tracking change only if it beats them at the transition and trajectory levels. A session-level gain alone is not enough, because it may come from knowing typical scores rather than from reading the dialogue. The no-change baseline predicts the same score at every session of a trajectory; its change MAE therefore equals the mean absolute true change. Metadata-only baselines use the label distribution, session position, and profile identity (Section~\ref{subsec:profile-audit}).

\section{Benchmark Validation}
\label{sec:analysis}

We validate three properties of the released data that the evaluation relies on; these checks do not establish clinical validity (Section~\ref{sec:limitations}).

\phantomsection\label{subsec:label-fidelity}
\paragraph{Label fidelity.}
The labels are self-consistent: the simulated clients' self-reports closely match the controlled states at the session, transition, and trajectory levels. Table~\ref{tab:corpus-statistics} reports the metrics of Section~\ref{subsec:protocol} with the self-report treated as a prediction of the controlled state. Across datasets, the self-report has a session MAE of $0.294$ to $0.541$ and a change MAE of $0.311$ to $0.626$. For large changes, it recovers the direction in $99.96\%$--$100\%$ of transitions, and $92.1\%$--$95.4\%$ of complete trajectories keep their trajectory type.

\begin{figure}[t]
  \centering
  \captionsetup{justification=centering}
  \begin{minipage}[t]{0.49\linewidth}
    \centering
    \includegraphics[width=\linewidth]{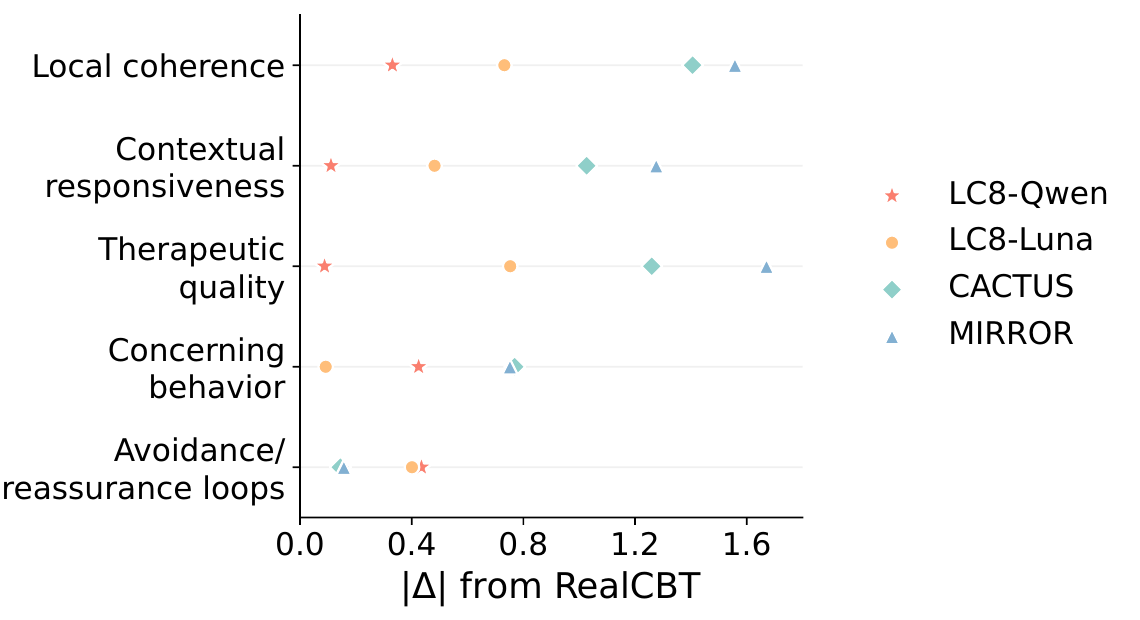}
    \caption{Rating differences from \realcbt.}
    \label{fig:snippet-audit}
  \end{minipage}\hfill
  \begin{minipage}[t]{0.49\linewidth}
    \centering
    \definecolor{hwSession}{HTML}{DB614F}
\definecolor{hwChange}{HTML}{F18C25}
\definecolor{hwDirection}{HTML}{6A9ACE}
\begin{tikzpicture}
  \pgfplotsset{
    hwaxis/.style={scale only axis, width=0.47\linewidth, height=0.38\linewidth,
      xmin=-0.4, xmax=3.4, xtick={0,1,2,3}, xticklabels={0,1,2,All},
      tick label style={font=\footnotesize}, label style={font=\footnotesize}},
    hwline/.style={line width=1.1pt, mark size=2.1pt,
      error bars/.cd, y dir=both, y explicit, error bar style={line width=0.6pt}},
  }
  \begin{axis}[hwaxis, axis y line*=left, axis x line*=bottom,
      ymin=2.55, ymax=4.4, ytick={3.0,3.5,4.0},
      yticklabel style={/pgf/number format/fixed, /pgf/number format/precision=1,
        /pgf/number format/fixed zerofill},
      xlabel={History window $k$}, ylabel={MAE $\downarrow$},
      ymajorgrids, grid style={black!12},
      legend style={draw=none, fill=none, font=\scriptsize, at={(0.02,0.98)},
        anchor=north west, row sep=-1pt}, legend cell align=left]
    \addplot[color=hwSession, mark=*, mark options={fill=hwSession, draw=white}, hwline]
      coordinates {(-0.06,3.594) += (0,0.145) -= (0,0.133)
                   (0.94,3.508) += (0,0.136) -= (0,0.139)
                   (1.94,3.503) += (0,0.137) -= (0,0.129)
                   (2.94,3.505) += (0,0.140) -= (0,0.140)};
    \addlegendentry{Session MAE $\downarrow$}
    \addplot[color=hwChange, mark=square*, mark options={fill=hwChange, draw=white}, hwline]
      coordinates {(-0.06,3.017) += (0,0.115) -= (0,0.115)
                   (0.94,2.920) += (0,0.121) -= (0,0.110)
                   (1.94,2.904) += (0,0.115) -= (0,0.113)
                   (2.94,2.834) += (0,0.118) -= (0,0.120)};
    \addlegendentry{Change MAE $\downarrow$}
    \addlegendimage{color=hwDirection, mark=triangle*, line width=1.1pt,
      mark options={fill=hwDirection, draw=white}}
    \addlegendentry{Direction error $\downarrow$}
  \end{axis}
  \begin{axis}[hwaxis, axis y line*=right, axis x line=none,
      ymin=0.07, ymax=0.625, ytick={0.1,0.2,0.3,0.4,0.5,0.6},
      yticklabel style={color=hwDirection, /pgf/number format/fixed,
        /pgf/number format/precision=1},
      ylabel={Direction error rate $\downarrow$}, ylabel style={color=hwDirection},
      y axis line style={color=hwDirection}, ytick style={color=hwDirection}]
    \addplot[color=hwDirection, mark=triangle*, mark options={fill=hwDirection, draw=white}, hwline]
      coordinates {(0.06,0.258) += (0,0.040) -= (0,0.042)
                   (1.06,0.307) += (0,0.045) -= (0,0.039)
                   (2.06,0.305) += (0,0.039) -= (0,0.039)
                   (3.06,0.317) += (0,0.041) -= (0,0.042)};
  \end{axis}
\end{tikzpicture}
    \caption{AIDA history-window sweep on \mbox{\datasetluna}.}
    \label{fig:history-window}
  \end{minipage}
\end{figure}

\phantomsection\label{subsec:plausibility}
\paragraph{Dialogue plausibility.}
The dialogues are close to real counseling: \datasetqwen and \datasetluna are closer to \realcbt~\citep{Wang2025FeelTD} than the earlier synthetic corpora CACTUS and MIRROR~\citep{lee2024cactus,kim2025mirror} in language, in rated behavior (Figure~\ref{fig:snippet-audit}), and in how hard they are to tell apart from it. Over seven language features, such as word count and the share of negative-emotion words, the distance from \realcbt is $0.114$ and $0.181$ for \datasetqwen and \datasetluna, compared with $0.274$ and $0.254$ for CACTUS and MIRROR, where lower is closer; \citet{wu2026emotrack} first reported all values except that of \datasetluna. A GPT-5.4 judge that is not told the source rates matched four- and eight-utterance snippets on five dimensions adapted from SIM-VAIL~\citep{weilnhammer2026simvail}; on four of them, both datasets differ less from \realcbt than CACTUS and MIRROR (Figure~\ref{fig:snippet-audit}). Asked to tell \realcbt snippets apart from synthetic ones, the judge reaches accuracy $0.832$ and $0.833$ for CACTUS and MIRROR but only $0.737$ and $0.688$ for \datasetqwen and \datasetluna. Appendix~\ref{sec:add_dataset_validation} gives the method details.

\phantomsection\label{subsec:profile-audit}
\paragraph{Severity information beyond metadata.}
The dialogue carries severity information that metadata does not: every transcript-based method beats the metadata-only baselines in session MAE, whereas knowing the client or the session position offers no shortcut. For each metric, we compare the best of five metadata-only baselines on the validation split (Appendix~\ref{sec:add_dataset_validation}) with the five transcript-based methods of Section~\ref{subsec:setup}. None of the baselines that use the profile improves on the global median (Table~\ref{tab:profile-audit}). Every transcript-based method, in contrast, achieves a lower session MAE; the best configuration reduces it from $3.903$ to $2.594$ (Table~\ref{tab:longcounsel-methods}).

\begin{table*}[ht]
  \centering
  \small
  \caption{Session MAE of metadata-only baselines with 95\% confidence intervals that resample complete profiles.}
  \label{tab:profile-audit}
  \begin{tabular}{lccc}
    \toprule
    \textbf{Metadata-only baseline}
      & \textbf{\datasetqwen}
      & \textbf{\datasetluna}
      & \makecell{\textbf{\datasetgpt}} \\
    \midrule

    Global median
      & 3.903 [3.703, 4.090]
      & 3.950 [3.763, 4.152]
      & 4.258 [3.714, 4.818] \\
    Same-profile median
      & 4.313 [4.076, 4.546]
      & 4.420 [4.207, 4.642]
      & -- \\
    Profile-by-session median
      & 4.468 [4.232, 4.702]
      & 4.520 [4.321, 4.733]
      & -- \\
    Profile + session ridge
      & 3.988 [3.834, 4.133]
      & 4.255 [4.092, 4.426]
      & 4.323 [3.817, 4.844] \\

    \bottomrule
  \end{tabular}
\end{table*}

\section{Experiment and Findings}
\label{sec:dataset-utility-evaluation}

In this section, we evaluate whether existing depression assessment methods can track changes in severity across counseling sessions. We first introduce the experimental setup (Section~\ref{subsec:setup}) and examine whether session-level accuracy translates into reliable longitudinal tracking (Section~\ref{subsec:longitudinal-results}). We then analyze clinically meaningful transitions (Section~\ref{subsec:transitions}), fluctuating trajectories (Section~\ref{subsec:worsening-trajectories}), and changes in individual symptoms (Section~\ref{subsec:symptom-change}). Finally, we investigate how access to earlier sessions affects change estimation and the identification of improvement or worsening (Section~\ref{subsec:history-utility}).

\subsection{Setup}
\label{subsec:setup}

\paragraph{Data and Metrics.}\datasetqwen provides the primary comparison under the protocol of Section~\ref{subsec:protocol}, across nine configurations spanning five existing methods; the session MAEs in Table~\ref{tab:longcounsel-methods} were first reported by~\citet{wu2026emotrack}. \datasetluna supports a replication, an AIDA history-window sweep, and a profile-disjoint evaluation. Appendix~\ref{sec:add_experiment_setting} provides method variants, refitting settings, and confidence-interval procedures. Five metrics introduced in Section~\ref{subsec:protocol} are used for evaluation.

\paragraph{Methods.}
We evaluate five reproducible methods with continuous severity outputs to test whether session-level depression assessment extends to longitudinal tracking.
They span structured LLM feature extraction with AIDA~\citep{lee2026interpretable}, questionnaire completion with LMIQ~\citep{rosenman2024llm}, symptom prediction~\citep{milintsevich2023towards}, and supervised transcript encoding, including the method of \citet{lau2023automatic} and EnsemBERT~\citep{ravenda2025transforming}.
History variants of AIDA and LMIQ test whether earlier sessions improve tracking, while Qwen3-Embedding variants of Lau et al.\ and EnsemBERT assess sensitivity to text representations, yielding nine configurations.
Appendix~\ref{sec:add_experiment_setting} provides implementation details.

\begin{table*}[ht]
  \centering
  \small
  \caption{Session- and transition-level results on \datasetqwen with 95\% confidence intervals.}
  \label{tab:longcounsel-methods}
  \begin{tabular}{lccc}
    \toprule
    \textbf{Method} & \makecell{\textbf{Session}\\\textbf{MAE $\downarrow$}}
      & \makecell{\textbf{Change MAE} \textbf{$\downarrow$}}
      & \makecell{\textbf{Direction}\\\textbf{accuracy $\uparrow$}} \\
    \midrule
    AIDA~\citep{lee2026interpretable} & 3.221 [3.108, 3.337]
      & 3.220 [3.102, 3.330] & 0.762 [0.724, 0.802] \\
    AIDA + history~\citep{lee2026interpretable} & 3.252 [3.136, 3.375]
      & 2.826 [2.718, 2.933] & 0.643 [0.602, 0.684] \\
    LMIQ~\citep{rosenman2024llm} & 2.979 [2.876, 3.088]
      & 3.082 [2.964, 3.201] & 0.779 [0.740, 0.815] \\
    LMIQ + history~\citep{rosenman2024llm} & 2.959 [2.840, 3.074]
      & 2.708 [2.601, 2.819] & 0.696 [0.659, 0.731] \\
    \citet{milintsevich2023towards}
      & 2.695 [2.605, 2.792] & 3.252 [3.117, 3.396] & 0.828 [0.797, 0.859] \\
    EnsemBERT~\citep{ravenda2025transforming} & 2.909 [2.796, 3.034] & 3.468 [3.318, 3.634] & 0.721 [0.680, 0.760] \\
    EnsemBERT (Qwen3-Emb)~\citep{ravenda2025transforming}
     & 2.749 [2.651, 2.851] & 3.098 [2.962, 3.241] & 0.753 [0.716, 0.789] \\
    \citet{lau2023automatic} & 2.620 [2.534, 2.709]
      & 3.078 [2.957, 3.194] & 0.822 [0.790, 0.854] \\
    Lau et al. (Qwen3-Emb)~\citep{lau2023automatic}
      & \textbf{2.594 [2.509, 2.682]} & 3.059 [2.944, 3.176]
      & \textbf{0.833 [0.801, 0.861]} \\
    \midrule
    \makecell[l]{Best dialogue-free baseline}
      & 3.903 [3.700, 4.111] & \textbf{2.478 [2.346, 2.607]} & 0.557 [0.519, 0.598] \\
    \bottomrule
  \end{tabular}
\end{table*}

\subsection{Session-level accuracy does not imply tracking}
\label{subsec:longitudinal-results}

\begin{table*}[ht]
  \centering
  \small
  \caption{Transition- and symptom-level results on \datasetqwen (five-seed means).}
  \label{tab:transition-symptom}
  \begin{tabular}{lcccccc}
    \toprule
    & \multicolumn{2}{c}{\makecell{\textbf{Direction}\\\textbf{accuracy $\uparrow$}}} & \multicolumn{2}{c}{\makecell{\textbf{Threshold}\\\textbf{detection $\uparrow$}}} & \multicolumn{2}{c}{\makecell{\textbf{Symptom-level}\\\textbf{change}}} \\
    \cmidrule(lr){2-3}\cmidrule(lr){4-5}\cmidrule(l){6-7}
    \textbf{Method} & Worsening & Improvement & Upward & Downward & MAE $\downarrow$ & Direction $\uparrow$ \\
    \midrule
    AIDA & 0.746 & 0.776 & 0.205 & 0.249 & -- & -- \\
    AIDA + history & 0.708 & 0.587 & 0.136 & 0.087 & -- & -- \\
    LMIQ & 0.768 & 0.788 & 0.222 & 0.254 & -- & -- \\
    LMIQ + history & 0.727 & 0.670 & 0.113 & 0.104 & -- & -- \\
    \citet{milintsevich2023towards} & 0.835 & 0.822 & 0.281 & 0.319 & 0.483 & \textbf{0.733} \\
    EnsemBERT & 0.721 & 0.720 & 0.287 & 0.321 & 0.506 & 0.389 \\
    \makecell[l]{EnsemBERT\\(Qwen3-Emb)} & 0.742 & 0.761 & 0.268 & 0.313 & \textbf{0.462} & 0.372 \\
    \citet{lau2023automatic} & \textbf{0.836} & 0.810 & \textbf{0.298} & 0.313 & -- & -- \\
    \makecell[l]{\citet{lau2023automatic}\\(Qwen3-Emb)} & \textbf{0.836} & \textbf{0.831} & 0.297 & \textbf{0.330} & -- & -- \\
    \bottomrule
  \end{tabular}
\end{table*}

\findingbox{Gains in session-level accuracy do not carry over to change: no method estimates change better than the no-change baseline.} Although every transcript-based configuration on \datasetqwen outperforms the metadata-only baselines in session MAE and direction accuracy (Table~\ref{tab:longcounsel-methods}), the best dialogue-free baseline reaches a change MAE of $2.478$, compared with $2.708$ for the best configuration, LMIQ with history.

The metrics also rank the methods differently. LMIQ with history has lower session MAE than AIDA ($2.959$ versus $3.221$), yet lower direction accuracy ($0.696$ versus $0.762$). Among transcript-based methods, LMIQ with history also leads on change MAE, but Lau et al.\ with Qwen3 embeddings reaches direction accuracy $0.833$ despite a higher change MAE of $3.059$. The same ranking difference appears on \datasetluna (Appendix Table~\ref{tab:prefix-evaluation}, $L=5$). It also holds on the profile-disjoint split of \datasetluna (Table~\ref{tab:luna-profile-disjoint}).

\begin{table*}[ht]
\centering
\small
\caption{Results on the profile-disjoint \datasetluna split with 95\% confidence intervals; within each column, orange marks better and blue worse values.}
\label{tab:luna-profile-disjoint}
\begin{tabular}{lccc}
\toprule
\textbf{Method} & \textbf{Session MAE $\downarrow$} & \textbf{Change MAE $\downarrow$} & \textbf{Direction $\uparrow$} \\
\midrule
AIDA & \cellcolor[rgb]{0.70040,0.80790,0.88776}3.370 [2.956, 3.826] & \cellcolor[rgb]{0.77699,0.85477,0.91170}2.991 [2.667, 3.329] & \cellcolor[rgb]{0.87690,0.86529,0.83813}0.737 [0.596, 0.852] \\
AIDA + history & \cellcolor[rgb]{0.60783,0.75294,0.85883}3.429 [2.980, 3.899] & \cellcolor[rgb]{0.98963,0.82097,0.64238}2.631 [2.283, 3.026] & \cellcolor[rgb]{0.67041,0.77907,0.85402}0.671 [0.546, 0.804] \\
LMIQ & \cellcolor[rgb]{0.80553,0.87032,0.92061}3.303 [2.916, 3.740] & \cellcolor[rgb]{0.90867,0.89871,0.87425}2.847 [2.518, 3.187] & \cellcolor[rgb]{0.60783,0.75294,0.85883}0.651 [0.529, 0.762] \\
LMIQ + history & \cellcolor[rgb]{0.98310,0.85268,0.71269}3.059 [2.712, 3.421] & \cellcolor[rgb]{1.00000,0.79608,0.58430}\textbf{2.582 [2.250, 2.929]} & \cellcolor[rgb]{0.65789,0.77384,0.85498}0.667 [0.576, 0.766] \\
Lau et al. & \cellcolor[rgb]{1.00000,0.79608,0.58430}\textbf{2.975 [2.646, 3.370]} & \cellcolor[rgb]{0.67427,0.79280,0.88008}3.069 [2.737, 3.430] & \cellcolor[rgb]{1.00000,0.79608,0.58430}\textbf{0.827 [0.717, 0.913]} \\
\bottomrule
\end{tabular}
\end{table*}

This disconnect between session-level accuracy and tracking can arise because a method can achieve low session MAE by estimating each client's overall severity without capturing changes between sessions. On \datasetqwen, $77\%$ of the variance of the controlled totals lies between trajectories, and adjacent sessions differ by only $2.48$ points on average, less than the session MAE of every method ($2.59$--$3.25$). An oracle that repeats each trajectory's mean controlled score at every session has a session MAE of $1.70$, lower than any method, yet estimates no change. An estimated change, in contrast, is accurate only when the errors of its two session estimates are similar, because its error is the difference between their errors. In Figure~\ref{fig:example}, EnsemBERT (Qwen3-Emb) underestimates S1--S4 by 2--3 points each, and these errors cancel in its changes, whereas AIDA + history remains near 12 from S2 onward while the controlled score moves.

\subsection{Clinically meaningful changes are often missed}
\label{subsec:transitions}

\findingbox{Methods recover the direction of most large changes but miss most crossings of the screening cutoff.} Without history, direction accuracy for large changes ranges from $0.72$ to $0.84$ on \datasetqwen and is similar for worsening and improvement (Table~\ref{tab:transition-symptom}). Threshold detection is far lower: no configuration detects more than $30\%$ of upward or $33\%$ of downward crossings on \datasetqwen, and at most $40\%$ of either on \datasetluna (Appendix Table~\ref{tab:transition-symptom-luna}). Crossings are hard to detect because a method must place both sessions on the correct side of the cutoff: two-thirds of the sessions at a crossing lie within three points of the cutoff, about one session MAE, and methods place each of them on the correct side only $53$--$64\%$ of the time. In Figure~\ref{fig:example}, EnsemBERT (Qwen3-Emb) places S4 above and S5 below the cutoff, whereas AIDA + history leaves S5 above it.

\subsection{Fluctuating trajectories are often missed}
\label{subsec:worsening-trajectories}

\begin{table*}[t]
  \centering
  \small
  \caption{Trajectory recovery on \datasetqwen.}
  \label{tab:trajectory-qwen}
  \begin{tabular}{lccc}
    \toprule
    & & \multicolumn{2}{c}{\textbf{Recall $\uparrow$}} \\
    \cmidrule(l){3-4}
    \textbf{Method} & \textbf{Macro-F1 $\uparrow$} & Fluctuating & Stable \\
    \midrule
    AIDA & 0.343 & 0.431 & 0.429 \\
    AIDA + history & 0.260 & 0.098 & 0.755 \\
    LMIQ & 0.447 & 0.410 & 0.634 \\
    LMIQ + history & 0.322 & 0.180 & 0.849 \\
    Milintsevich et al. & 0.487 & 0.466 & 0.626 \\
    EnsemBERT & 0.465 & 0.469 & 0.510 \\
    EnsemBERT (Qwen3-Emb) & 0.479 & 0.436 & 0.622 \\
    Lau et al. & \textbf{0.507} & \textbf{0.484} & 0.631 \\
    Lau et al. (Qwen3-Emb) & 0.505 & 0.480 & 0.642 \\
    \bottomrule
  \end{tabular}
\end{table*}

\findingbox{All configurations recover about half or fewer of fluctuating trajectories, and history turns most of them into stable predictions.} Table~\ref{tab:trajectory-qwen} applies the four-type rule to the predicted scores. The recall of fluctuating trajectories is at most $0.484$, and for LMIQ, history improves change MAE but lowers this recall from $0.410$ to $0.180$ while raising stable recall from $0.634$ to $0.849$; $69.5\%$ of fluctuating trajectories are then predicted as stable (Appendix~\ref{apdx:trajectory-recovery}). This failure is visible only with labels throughout the trajectory, because a fluctuating trajectory is defined by its intermediate sessions. Stratified MAE shows that higher severity partly explains the higher session error on worsening trajectories reported by~\citet{wu2026emotrack} (Appendix~\ref{apdx:trajectory-recovery}).

\subsection{Symptom-level changes are not identified}
\label{subsec:symptom-change}

\findingbox{A correct total-score direction can arise without identifying which symptoms changed.} In Table~\ref{tab:transition-symptom}, the three configurations that output symptom scores all have larger symptom-level change error than the no-change baseline ($0.46$--$0.51$ versus $0.31$), and \datasetluna shows the same ($0.42$--$0.47$ versus $0.30$). For symptom changes of at least one point, \citet{milintsevich2023towards} recover the direction in $73\%$ of cases, whereas both EnsemBERT variants recover $37$--$39\%$, below their total-score direction accuracy of $0.72$--$0.75$; EnsemBERT predicts discrete symptom scores and leaves about two-thirds of them unchanged between sessions.

\subsection{History improves magnitude but weakens direction}
\label{subsec:history-utility}

\findingbox{Adding history lowers change error but reduces direction accuracy and threshold detection, mostly because improvements are missed.} On \datasetqwen, adding history lowers improvement accuracy from $0.776$ to $0.587$ for AIDA and from $0.788$ to $0.670$ for LMIQ, while worsening accuracy falls by only about $0.04$; \datasetluna shows the same asymmetry (Appendix Table~\ref{tab:transition-symptom-luna}). An AIDA history-window sweep on \datasetluna shows that reading more history lowers change MAE but also lowers direction accuracy (Figure~\ref{fig:history-window}). History also reduces threshold detection by $34$--$65\%$ (Table~\ref{tab:transition-symptom}). History lowers change MAE by shrinking the estimated changes, not by following them better: their standard deviation falls from $3.4$ to $2.2$ for AIDA and from $3.9$ to $2.3$ for LMIQ (true changes: $3.6$), and their correlation with the true changes falls from $0.31$ to $0.23$ and from $0.39$ to $0.30$. It also shrinks estimated improvements far more than estimated worsening: on large improvements of about seven points on \datasetqwen, history shrinks the average estimated change from $-2.3$ to $-0.4$ points for AIDA and from $-3.4$ to $-1.2$ for LMIQ, but on large worsening only from $+2.0$ to $+1.5$ and from $+3.1$ to $+1.7$. This suggests that symptoms described in earlier sessions keep the estimate high after the client improves, as at S5 in Figure~\ref{fig:example}. This extends the observation of~\citet{wu2026emotrack}, that previous-session memory can increase error after abrupt changes, to existing methods and to direction accuracy.

\section{Limitations}
\label{sec:limitations}

\paragraph{Label validity.} Although the controlled states and simulated self-reports are consistent, our data remain synthetic, and this agreement does not establish that the generated dialogues faithfully represent real clients with the same symptom severity. The self-reports are produced by the same client simulator using the same private cues that guide dialogue generation.

\paragraph{Self-report bias at higher severity.} For sessions with a controlled PHQ-8 total of at least 15, self-reports have larger errors and are biased toward greater severity (Appendix~\ref{sec:add_dataset_validation}). Evaluation against the controlled state is unaffected, but models trained on self-reports may inherit this bias.

\paragraph{Scope of empirical grounding.} Although each component is grounded in real data, the information combined in a synthetic trajectory comes from different individuals: the construction protocol of \counselarc~\citep{wu2026emotrack} pairs profiles derived from real counseling cases with longitudinal PHQ-9 trajectories from separate empirical resources. Jointly collected longitudinal counseling data or expert annotations would allow future evaluation of profile--state compatibility and symptom-level expression.

\section{Conclusion}
\label{sec:conclusion}

We presented \longcounsel, a benchmark for tracking depression severity across counseling sessions. None of the five evaluated methods outperforms the no-change baseline in change estimation, and important transitions, trajectory patterns, and symptom changes are often missed. Earlier-session context can further obscure improvement by shrinking estimated changes. These findings motivate methods that estimate change directly and use history without flattening real changes; \longcounsel provides the labels and metrics to evaluate this progress.

\begin{ack}
We thank Adrian Toh, Clinical Psychologist, for his valuable clinical insights and expert guidance in clinical psychology, particularly in informing the design and development of the counseling dataset.

This research is supported by the National Research Foundation, Singapore and Infocomm Media Development Authority under its Trust Tech Funding Initiative. Any opinions, findings and conclusions or recommendations expressed in this material are those of the authors and do not reflect the views of National Research Foundation, Singapore and Infocomm Media Development Authority. This research is also in part supported by Zhaomin Wu's National Research Foundation (NRF) Postdoctoral Award.
\end{ack}

\bibliographystyle{plainnat}
\bibliography{references}

\clearpage
\addcontentsline{toc}{section}{Appendix}
\noindent\begin{minipage}{\linewidth}
\part{Appendix}
\parttoc
\end{minipage}\par

\appendix

\section{Expanded Survey of Counseling Dialogue Resources}
\label{apdx:dataset-survey}

Table~\ref{tab:counseling-datasets-expanded} complements the comparison in Table~\ref{tab:evaluation-gap} with a broader inventory of counseling, psychotherapy, and motivational-interviewing dialogue resources. We include corpora that contain full counselor--client exchanges or explicitly construct complete therapy sessions; general mental-health question answering, social-media prediction, and counselor-knowledge benchmarks fall outside this scope. To avoid confusing a long conversation with care over time, we use \emph{multi-session counseling dialogues} only for separate sessions linked to the same client or simulated persona. Likewise, a resource is \emph{PHQ-supervised} only if each session has a PHQ score, rather than a general depression topic, a dialogue-quality rating, or another outcome measure.

\paragraph{Human-authored and source-derived dialogue.}
KokoroChat collects one-session role plays by trained counselors, whereas HOPE, MEMO, \realcbt, AnnoMI, and MIDAS transcribe or annotate counseling demonstrations from public media~\citep{Qi2025KokoroChatAJ,Malhotra2021SpeakerAT,Srivastava2022CounselingSU,Wang2025FeelTD,Wu2022AnnoMIAD,Gunal2025ExaminingSC}. The licensed Counseling and Psychotherapy Transcripts collection includes therapy material spanning multiple encounters~\citep{CPTFullCollection}. These resources provide valuable human dialogue, but do not pair every session with a standardized depression score. \daic is the only PHQ-supervised resource in this group, although each participant contributes one semi-structured assessment interview rather than a series of counseling sessions~\citep{Gratch2014TheDA,burdisso2024daic}.

\paragraph{Generated and reconstructed dialogue.}
PsyDT, PsyDial, and CPsyCoun use source conversations, long-form counseling material, or case reports to reconstruct privacy-preserving multi-turn sessions~\citep{xie2025psydt,qiu2025psydial,zhang2024cpsycoun}. CACTUS, HamRaz, MIRROR, KMI, and Thousand Voices of Trauma instead generate sessions around particular therapeutic frameworks, cultures, modalities, or disorders~\citep{lee2024cactus,abbasi2025hamraz,kim2025mirror,kim2025kmi,suhas2025thousand}. TheraPhase, MusPsy, and PsychEval explicitly organize records or generated conversations across treatment stages and multiple sessions~\citep{na2026theraphase,wang2026muspsy,pan2026psycheval}, but none labels each session with a standardized PHQ score. Among the surveyed resources, only \counselarc~\citep{wu2026emotrack} and \longcounsel provide these labels; Table~\ref{tab:evaluation-gap} compares how the resources are evaluated.

\begin{table*}[t]
  \centering
  \small
  \caption{Expanded inventory of counseling and motivational-interviewing dialogue resources.}
  \label{tab:counseling-datasets-expanded}
  \begin{tabular}{l l c c c}
    \toprule
    \textbf{Dataset} &
    \textbf{Dialogue source} &
    \textbf{Lang.} &
    \makecell{\textbf{Multi-session}\\\textbf{dialogues}} &
    \makecell{\textbf{PHQ-}\\\textbf{supervised}} \\
    \midrule
    \multicolumn{5}{@{}l}{\textit{Human-authored or human-source-derived dialogue}} \\
    \midrule
    KokoroChat~\citep{Qi2025KokoroChatAJ} & Human role-play & JA & \xmark & \xmark \\
    HOPE~\citep{Malhotra2021SpeakerAT} & Public video & EN & \xmark & \xmark \\
    MEMO~\citep{Srivastava2022CounselingSU} & Public video & EN & \xmark & \xmark \\
    \makecell[l]{Counseling and Psychotherapy\\Transcripts~\citep{CPTFullCollection}} & Licensed transcripts & EN & \cmark & \xmark \\
    \realcbt~\citep{Wang2025FeelTD} & Public video & EN & \xmark & \xmark \\
    AnnoMI~\citep{Wu2022AnnoMIAD} & Public video & EN & \xmark & \xmark \\
    MIDAS~\citep{Gunal2025ExaminingSC} & Public video & ES & \xmark & \xmark \\
    \makecell[l]{\daic\\\citep{Gratch2014TheDA,burdisso2024daic}} & Clinical interview & EN & \xmark & \cmark \\
    \midrule
    \multicolumn{5}{@{}l}{\textit{LLM-generated or LLM-reconstructed dialogue}} \\
    \midrule
    PsyDT / PsyDTCorpus~\citep{xie2025psydt} & LLM reconstruction & ZH & \xmark & \xmark \\
    PsyDial~\citep{qiu2025psydial} & LLM reconstruction & ZH & \xmark & \xmark \\
    CACTUS~\citep{lee2024cactus} & LLM generation & EN & \xmark & \xmark \\
    \makecell[l]{CPsyCoun / Memo2Demo\\\citep{zhang2024cpsycoun}} & LLM reconstruction & ZH & \xmark & \xmark \\
    HamRaz~\citep{abbasi2025hamraz} & Hybrid LLM generation & FA & \xmark & \xmark \\
    MIRROR~\citep{kim2025mirror} & LLM generation & EN & \xmark & \xmark \\
    KMI~\citep{kim2025kmi} & LLM generation & KO & \xmark & \xmark \\
    \makecell[l]{Thousand Voices of Trauma\\\citep{suhas2025thousand}} & LLM generation & EN & \xmark & \xmark \\
    TheraPhase~\citep{na2026theraphase} & Case-grounded generation & ZH & \cmark & \xmark \\
    MusPsy~\citep{wang2026muspsy} & Case-grounded generation & ZH & \cmark & \xmark \\
    PsychEval~\citep{pan2026psycheval} & Case-grounded generation & ZH & \cmark & \xmark \\
    \midrule
    \counselarc~\citep{wu2026emotrack} & Case-grounded generation & EN & \cmark & \cmark \\
    \longcounsel (this work) & Case-grounded generation & EN & \cmark & \cmark \\
    \bottomrule
  \end{tabular}
\end{table*}

\section{Benchmark Construction Details}
\label{apdx:construction}

This appendix complements Section~\ref{sec:construction}. Construction steps inherited from~\citet{wu2026emotrack}, including client-profile conversion, trajectory reconstruction, cue pruning, and the prompt templates, are documented there and in the supplementary code; we report only the settings specific to this benchmark. Each self-report is the average of five PHQ-8 questionnaires that the simulated client fills out after the session, with the questions in shuffled order~\citep{wu2026emotrack}.

\subsection{Quality control and regeneration}
\label{apdx:regeneration}

The checks follow~\citet{wu2026emotrack}. Quality checks reject empty, non-English, numeric-only, label-leaking, or excessively repeated turns, and each self-report must answer all eight questions exactly once with verbatim option text. Generation allows up to three counselor attempts, five client attempts, and three attempts per self-report sample, with network retries and JSON repair handled within each attempt. A generated run is complete only when its transcript is nonempty and parseable and all five sessions have a controlled state and self-report samples. Sessions that still fail or are interrupted are regenerated until the run is complete.

\subsection{Released data schema}
\label{apdx:schema}

The core release organizes each trajectory under its \texttt{run\_id}. Table~\ref{tab:release-schema} summarizes the three core files.

\begin{table*}[htbp]
  \centering
  \small
  \caption{Core files released for each trajectory.}
  \label{tab:release-schema}
  \begin{tabularx}{\linewidth}{>{\raggedright\arraybackslash\hsize=.8\hsize}X
    >{\raggedright\arraybackslash\hsize=1.2\hsize}X}
    \toprule
    \textbf{File} & \textbf{Contents} \\
    \midrule
    \texttt{transcript.jsonl} & One row per utterance: one-based session and exchange indices, speaker, visible text, unique message id. Each session has 20 counselor--client exchanges, or 40 rows. \\
\midrule
    \texttt{ground\_truth.json} & Five session records containing the controlled PHQ-8 symptom vector, total in $[0,24]$, and severity band. This is the authoritative evaluation target. \\\midrule
    \texttt{self\_report.json} & Five session records containing five raw questionnaire samples and their symptom-wise aggregate. Aggregate symptom scores and totals are numeric and need not be integers. \\
    \bottomrule
  \end{tabularx}
\end{table*}

The transcript speakers are \texttt{Counselor} and \texttt{Client}; the visible utterance is stored in \texttt{text}. Benchmark loaders render only speaker-tagged visible text. The client-side \texttt{emotion} response is not written to the release transcript. Profile text, behavior cues, session focus, disclosure plan, summaries, memories, and carryover notes are used during generation and are not included in the released trajectories. Dataset, generator, and split membership are given in separate manifest files. A complete trajectory contains five sessions and 200 transcript rows.

\subsection{Generator-specific settings}
\label{apdx:sampling}

For dialogue generation, \datasetqwen uses temperature $T=0.6$, top-$p$ $0.95$, top-$k$ $20$, a 512-token cap per turn, and a 4{,}096-token reasoning budget; \datasetluna and \datasetgpt use the API default decoding, a 1{,}536-token cap per turn, and the low reasoning setting. GPT-5.4-mini and GPT-5.6 Luna use the Responses API path, whose adapter omits temperature; both therefore use the API's fixed/default sampling. All random assignments use seed $42$; the local Qwen server additionally uses seed $0$. These seeds make the assignments and the choice of turn-form templates reproducible, but sampled text is not guaranteed to be identical across reruns.

For \datasetqwen, session-summary and self-report calls set temperature $0$. For \datasetgpt and \datasetluna, these calls omit temperature. Summary outputs are capped at 1024 tokens. Self-report outputs are capped at 1024 tokens for \datasetqwen and 1536 for the GPT-5 datasets.

\subsection{Trajectory types}
\label{apdx:trajectory-types}

We classify the trajectories into four types using their controlled PHQ-8 totals. An \emph{improving} trajectory has a final score at least five points below its starting score, while a \emph{worsening} trajectory ends at least five points higher. A \emph{fluctuating} trajectory has endpoints that differ by less than five points but an internal range of at least five points. All remaining trajectories are \emph{stable}. The five-point threshold is the large-change criterion of Section~\ref{subsec:protocol}.

\subsection{Split manifests}
\label{apdx:splits}

Each dataset has a seed-42 trajectory manifest. The regular \datasetqwen and \datasetluna manifests contain $2{,}583/369/738$ trajectories and allow the same profile to appear in train, validation, and test because each profile is combined with multiple trajectories. The \datasetgpt manifest contains $258/37/74$ trajectories and is profile-disjoint because it contains one trajectory per profile. The manifests store each \texttt{run\_id} and its split explicitly; the released splits therefore do not depend on filename order. 

\section{Benchmark Validation Details}
\label{sec:add_dataset_validation}

This appendix details the language, label-fidelity, and metadata-only baseline analyses of Section~\ref{sec:analysis}.

The seven language features are sentence count, word count, words per sentence, Flesch reading ease, and the proportions of positive-emotion, negative-emotion, and absolutist words (such as \emph{always} and \emph{never}). For each feature and speaker role, the normalized quantile distance is the mean absolute gap from \realcbt at the 10th, 25th, 50th, 75th, and 90th percentiles, divided by a fixed scale pooled from \realcbt, \datasetqwen, CACTUS, and MIRROR. This scale is the 90th--10th percentile range, with standard deviation used when that range is zero. The corpus-level distance averages all 14 feature--role distances, and each corpus contributes $7{,}450$ utterances after adjacent same-speaker turns are merged. Snippet ratings range from 1 to 10; higher scores indicate greater coherence, responsiveness, and therapeutic quality, but more concerning behavior and avoidance or reassurance loops.

\paragraph{Multidimensional snippet outcomes.}
Table~\ref{tab:snippet-multidimensional} reports the pooled means, effect sizes, and equivalence decisions behind three of the five dimensions in Figure~\ref{fig:snippet-audit}: therapeutic quality and the two local-safety dimensions, concerning behavior and avoidance or reassurance loops. The table covers the two LC8 datasets, for which equivalence against \realcbt was tested; local coherence, contextual responsiveness, and the CACTUS and MIRROR comparisons are reported descriptively in Figure~\ref{fig:snippet-audit}. Continuous ratings use ordinary least squares (OLS) with source-cluster-robust covariance and adjust for log token count, position, first speaker, and, in pooled models, snippet length. Effect sizes are small-sample-corrected Hedges $g$. Equivalence uses two one-sided tests (TOST) with standardized margin $|d|<0.30$ and Holm correction across the two synthetic comparisons within each outcome and scope. The separate source-identification task reports balanced accuracy with 2,000 cluster-bootstrap samples.

\begin{table*}[ht]
  \centering
  \small
  \caption{Selected pooled outcomes from the multidimensional snippet comparison.}
  \label{tab:snippet-multidimensional}
  \begin{tabular}{lccc}
    \toprule
    \textbf{Outcome} & \textbf{\realcbt mean}
      & \makecell{\textbf{\datasetqwen}\\\textbf{mean / $g$ / Eq}}
      & \makecell{\textbf{\datasetluna}\\\textbf{mean / $g$ / Eq}} \\
    \midrule
    Therapeutic quality $\uparrow$ & 6.190 & 6.102 / $-0.047$ / Yes
      & 6.943 / $+0.575$ / No \\
    Concerning behavior $\downarrow$ & 1.856 & 2.281 / $+0.420$ / No
      & 1.764 / $-0.083$ / Yes \\
    Avoidance/reassurance loops $\downarrow$ & 1.185 & 1.620 / $+0.515$ / No
      & 1.586 / $+0.528$ / No \\
    \bottomrule
  \end{tabular}
\end{table*}

\paragraph{Longitudinal label fidelity.}
Table~\ref{tab:dataset-validation} reports the per-dataset results summarized in Section~\ref{subsec:label-fidelity}. We also evaluate symptom exact match, symptom MAE, total-score bias (self-report minus controlled state), controlled--self-report correlation, and severity-band performance. Dataset-level 95\% intervals use 2,000 trajectory resamples with seed 20260825. The controlled-total-$\geq15$ values in Table~\ref{tab:dataset-validation} pool sessions from the 15--19 and 20--24 bands. In these sessions, self-report MAE rises to $3.382$ for \datasetqwen, $1.331$ for \datasetluna, and $1.558$ for \datasetgpt, above the respective overall averages; in \datasetqwen, this range contains $7.6\%$ of sessions, and every error shifts the self-report toward greater severity. Table~\ref{tab:label-diagnostics} reports additional symptom-level diagnostics.

\begin{table*}[ht]
  \centering
  \small
  \caption{Per-dataset label-fidelity results; brackets show 95\%
  trajectory-bootstrap confidence intervals.}
  \label{tab:dataset-validation}
  \begin{tabular}{lccc}
    \toprule
    \textbf{Metric}
      & \textbf{\datasetqwen}
      & \textbf{\datasetluna}
      & \textbf{\datasetgpt} \\
    \midrule

    Session MAE $\downarrow$
      & 0.521 [0.500, 0.543]
      & 0.294 [0.284, 0.306]
      & 0.541 [0.496, 0.589] \\
    Change MAE $\downarrow$
      & 0.626 [0.603, 0.648]
      & 0.311 [0.300, 0.322]
      & 0.434 [0.401, 0.469] \\
    \makecell[l]{Trajectory-type agreement $\uparrow$}
      & 0.921 [0.912, 0.929]
      & 0.954 [0.947, 0.961]
      & 0.946 [0.921, 0.967] \\

    \midrule
    MAE, controlled total $\geq 15$ $\downarrow$
      & 3.382 [3.305, 3.457]
      & 1.331 [1.277, 1.388]
      & 1.558 [1.359, 1.756] \\

    \bottomrule
  \end{tabular}
\end{table*}

\begin{table*}[ht]
  \centering
  \small
  \caption{Detailed label-fidelity diagnostics.}
  \label{tab:label-diagnostics}
  \begin{tabular}{>{\raggedright\arraybackslash}p{0.29\linewidth}
    ccc}
    \toprule
    \textbf{Metric} & \textbf{\datasetqwen} & \textbf{\datasetluna}
      & \textbf{\datasetgpt} \\
    \midrule
    Symptom exact match & 0.826 [0.822, 0.830] & 0.898 [0.895, 0.901]
      & 0.836 [0.824, 0.848] \\
    Symptom MAE & 0.069 [0.066, 0.071] & 0.037 [0.036, 0.038]
      & 0.068 [0.063, 0.074] \\
    Total-score bias & 0.133 [0.108, 0.160] & 0.293 [0.282, 0.304]
      & 0.538 [0.491, 0.587] \\
    Controlled--self-report correlation & 0.990 [0.990, 0.991]
      & 0.997 [0.997, 0.997] & 0.996 [0.995, 0.996] \\
    \bottomrule
  \end{tabular}
\end{table*}

\subsection{Metadata-only baselines}

Profile identity and session position are the only signals available when the dialogue and earlier PHQ labels are withheld. Five metadata-only baselines test how much these signals predict. The \emph{global median} predicts the median training self-report total for every session and represents the label distribution without profile or session information. The \emph{session-index mean} predicts, for each session position $t=1,\ldots,5$, the mean training self-report total at that position. The \emph{same-profile median} tests whether a stable profile identity predicts severity, while the \emph{profile-by-session median} additionally tests whether that association changes with session position. A regularized \emph{profile-plus-session ridge} model learns a joint combination of the two signals.

All five baselines are fitted on training self-reports and selected on validation self-reports before evaluation against test controlled totals. The baseline selected on the validation split is therefore the strongest prespecified metadata-only baseline for each metric. For session and change MAE, validation selects the global median. For large-change direction accuracy, it selects the session-index mean: a constant prediction never changes across sessions, and because Section~\ref{subsec:protocol} counts a predicted change of zero as an error, the global median has zero direction accuracy. The same protocol is applied to the fixed \datasetqwen split used by the transcript-based benchmark.

Table~\ref{tab:profile-audit} reports session MAE for the global median and the baselines that use the profile. None of these baselines improves on the global median in \datasetqwen or \datasetluna, and the naturally profile-disjoint \datasetgpt split shows the same pattern. Profile identity and session position therefore do not improve session MAE under the tested baselines.

\subsection{Comparison with transcript-based methods}

The transcript comparison asks whether the dialogue provides recoverable severity information beyond what the metadata-only baselines capture. It uses the five methods selected in Section~\ref{subsec:setup}; Appendix~\ref{sec:add_experiment_setting} documents their implementation and hyperparameters.

All methods use the fixed \datasetqwen split of $2{,}583/369/738$ train, validation, and test trajectories. Learned methods and the validation-selected metadata-only baseline follow the training and evaluation protocol of Section~\ref{subsec:protocol}; the baseline receives no dialogue. This comparison isolates the information contributed by transcript content under a shared target and split.

The best metadata-only baseline row in Table~\ref{tab:longcounsel-methods} combines baselines selected separately for each metric.

\section{Experiment Details and Additional Results}
\label{sec:add_experiment_setting}

\paragraph{Input features.}
Utterance embeddings come from frozen Qwen3-Embedding-8B~\citep{qwen3embedding8b}; the Qwen replacement uses last-token pooling, maximum length 256, and bfloat16. AIDA clinical features are extracted by Qwen3.5-35B-A3B~\citep{qwen35_35ba3b} using structured rubrics, and cached features are seed-matched with downstream training.

\paragraph{Splits and targets.}
The full \datasetqwen and \datasetluna comparisons report five training seeds, numbered 0--4. Training, selection, and evaluation follow Section~\ref{subsec:protocol}. The metadata-only baselines are deterministic and are reported without seed averaging.

The profile-disjoint \datasetluna evaluation uses 369 trajectories sampled uniformly without replacement with seed 42 and a fixed seed-42 profile-group assignment for $258/37/74$ train, validation, and test trajectories. The splits contain $166/27/51$ profiles with zero cross-split overlap. Each configuration uses five training seeds. The test set contains 370 sessions, 296 adjacent transitions, and 51 large-change transitions. Point estimates average seed-level metrics rather than predictions.

\paragraph{Shared method inputs.}
We implement AIDA~\citep{lee2026interpretable}, LMIQ~\citep{rosenman2024llm}, \citet{lau2023automatic}, \citet{milintsevich2023towards}, and EnsemBERT~\citep{ravenda2025transforming} from released algorithms or author code where available. History variants concatenate sessions 1 through the current session; other variants read only the current session. Models and checkpoints are selected on the validation split.

\paragraph{AIDA and LMIQ.}
AIDA extracts 23 structured features with Qwen3.5-35B-A3B, standardizes them, adds an intercept, and fits an ordinary least-squares linear head. Feature extraction uses a fixed inference seed, temperature 0, at most three parse attempts per question, and a transcript-hash cache. With extraction seed 0, the history-window sweep re-extracts features and refits this head separately for $k=0,1,2,$ and all available previous sessions, using combined training and validation self-reports for the final fit; larger windows nearly coincide with all available history. Because session 1 receives identical input under every window, the sweep scores all three metrics on sessions 2--5. LMIQ applies its released question bank and fits a random forest. Validation mean squared error (MSE) selects among 100, 200, or 300 trees and maximum depths 10, 20, or 30, with validation MAE and parameter order used for tie breaking.

\paragraph{Encoder-based methods.}
The standard Lau et al.\ dual encoder uses \texttt{all-mpnet-base-v2} with mean pooling; its Qwen replacement uses Qwen3-Embedding-8B with last-token pooling. Both use the current session, batch size 2, learning rate $3\times10^{-4}$, weight decay $0.01$, at most 200 epochs, patience 20, and validation-loss checkpoint selection. The Milintsevich et al.\ implementation uses the current session, 100 iterations, batch size 48, and patience 10. EnsemBERT uses client utterances, at most 30 posts with a two-word minimum, a 256-unit layer with two heads, learning rate $10^{-3}$, weight decay $10^{-4}$, batch size 64, at most 50 epochs, and patience 8; output, soft, and metadata losses have equal weight, and the metadata head supplies the prediction. Its standard encoder is \texttt{all-mpnet-base-v2}; the reported Qwen variant uses the frozen replacement described above.

\paragraph{Longitudinal metrics.}
Metrics follow Section~\ref{subsec:protocol}. They use saved predictions without additional rounding or clipping. For five-seed comparisons, each seed is scored first and the resulting metrics are averaged.

Trajectory types follow the rule in Appendix~\ref{apdx:trajectory-types}, applied to the controlled totals. For the severity-matched comparison, each worsening session is compared with sessions of the same controlled total from the comparison type, using seed-averaged absolute errors weighted by the worsening sessions' score distribution; totals with fewer than three comparison sessions are skipped, which keeps $92\%$ of worsening sessions for the comparison with stable trajectories. We also apply the same rule to predicted five-session score sequences to evaluate trajectory recovery, using macro-F1 over the four types and recall within each type.

On the profile-disjoint \datasetluna split, the session-index mean (Appendix~\ref{sec:add_dataset_validation}) serves as the metadata-only baseline. Its intervals use 2,000 complete-trajectory bootstrap samples with seed 20260828. The same trajectory draws are shared across methods and seeds, and paired intervals use within-replicate differences. The full \datasetqwen and \datasetluna longitudinal analyses use 2,000 trajectory resamples with seed 20260825. Cross-dataset intervals are unpaired because dataset assignments are independent.

\paragraph{Recovery of complete trajectories.}
\label{apdx:trajectory-recovery}
Trajectory recovery (Section~\ref{subsec:worsening-trajectories}) uses the nine transcript configurations from Table~\ref{tab:longcounsel-methods}, each with seeds 0--4. All configurations and seeds have predictions for the same 738 complete test trajectories on \datasetqwen and \datasetluna. We derive trajectory types from the saved floating-point predictions without further fitting or inference. For type $c$, recall is $\mathrm{TP}_c/N_c$, where $N_c$ is the number of controlled trajectories of that type and $\mathrm{TP}_c$ counts correct predictions. Macro-F1 averages $2\mathrm{TP}_c/(2\mathrm{TP}_c+\mathrm{FP}_c+\mathrm{FN}_c)$ equally over the four types, with $\mathrm{FP}_c$ and $\mathrm{FN}_c$ denoting false positives and false negatives.

Table~\ref{tab:trajectory-recovery} reports the resulting trajectory recovery. The \datasetqwen test set contains 93 improving, 58 worsening, 229 fluctuating, and 358 stable trajectories; the \datasetluna counts are 79, 61, 219, and 379. On \datasetqwen, adding history raises the proportion of fluctuating trajectories predicted as stable from $40.7\%$ to $69.5\%$ for LMIQ and from $37.5\%$ to $73.4\%$ for AIDA. \datasetluna shows the same loss of fluctuation recall with history for both methods. This identifies a concrete failure to recover intermediate changes despite improved average change error.

In the results of~\citet{wu2026emotrack} on \datasetqwen, broken down by trajectory type, every method has its lowest session MAE on stable and its highest on worsening trajectories, a gap of $0.67$--$1.24$ points. Matching sessions on the controlled total narrows the gap to stable sessions to $0.18$--$0.32$. The reduced gap indicates that higher symptom severity partly accounts for the original difference. The higher error also does not extend to recognizing the worsening type: eight of the nine \datasetqwen configurations recall worsening trajectories at least as often as improving ones.

\begin{table*}[ht]
  \centering
  \small
  \caption{Five-session trajectory recovery.}
  \label{tab:trajectory-recovery}
  \begin{tabular}{lccccc}
    \toprule
    \textbf{Method} & \textbf{Macro-F1} & \multicolumn{4}{c}{\textbf{Recall}} \\
    \cmidrule(lr){3-6}
     & & Improving & Worsening & Fluctuating & Stable \\
    \midrule
    \multicolumn{6}{l}{\datasetqwen ($n=738$)} \\
    AIDA & 0.343 & 0.256 & 0.241 & 0.431 & 0.429 \\
    AIDA + history & 0.260 & 0.037 & 0.293 & 0.098 & 0.755 \\
    LMIQ & 0.447 & 0.331 & 0.393 & 0.410 & 0.634 \\
    LMIQ + history & 0.322 & 0.069 & 0.255 & 0.180 & 0.849 \\
    Milintsevich et al. & 0.487 & 0.404 & 0.503 & 0.466 & 0.626 \\
    EnsemBERT & 0.465 & 0.465 & 0.534 & 0.469 & 0.510 \\
    EnsemBERT (Qwen3-Emb) & 0.479 & 0.430 & 0.476 & 0.436 & 0.622 \\
    Lau et al. & 0.507 & 0.447 & 0.514 & 0.484 & 0.631 \\
    Lau et al. (Qwen3-Emb) & 0.505 & 0.454 & 0.486 & 0.480 & 0.642 \\
    \midrule
    \multicolumn{6}{l}{\datasetluna ($n=738$)} \\
    AIDA & 0.378 & 0.352 & 0.184 & 0.342 & 0.611 \\
    AIDA + history & 0.247 & 0.013 & 0.207 & 0.053 & 0.885 \\
    LMIQ & 0.401 & 0.286 & 0.177 & 0.421 & 0.672 \\
    LMIQ + history & 0.313 & 0.081 & 0.151 & 0.167 & 0.892 \\
    Milintsevich et al. & 0.480 & 0.529 & 0.308 & 0.540 & 0.600 \\
    EnsemBERT & 0.449 & 0.435 & 0.390 & 0.400 & 0.626 \\
    EnsemBERT (Qwen3-Emb) & 0.436 & 0.385 & 0.302 & 0.404 & 0.684 \\
    Lau et al. & 0.493 & 0.491 & 0.331 & 0.484 & 0.668 \\
    Lau et al. (Qwen3-Emb) & 0.491 & 0.499 & 0.331 & 0.509 & 0.636 \\
    \bottomrule
  \end{tabular}
\end{table*}

\paragraph{Evaluation over longer trajectory prefixes.}
\label{apdx:prefix-evaluation}
We evaluate prefixes of length $L=2,3,4,5$ using the same complete trajectories, configurations, and seeds at every length. Session MAE covers sessions 1 through $L$; change MAE covers the $L-1$ adjacent transitions, and direction accuracy uses the eligible large-change transitions. The number of eligible large changes is 130, 260, 384, and 497 on \datasetqwen and 125, 259, 376, and 489 on \datasetluna. Prefix length changes the evaluated portion of the saved predictions, while the model and its history-input setting remain fixed.

Table~\ref{tab:prefix-evaluation} shows how the three metrics change as evaluation includes later sessions. All nine \datasetqwen configurations have lower direction accuracy over five sessions than over the first two, while average change error falls for eight configurations. Direction accuracy does not decrease at every intermediate prefix, and on \datasetluna it falls for every configuration except AIDA with history.

\begin{table*}[ht]
  \centering
  \small
  \caption{Session MAE/change MAE/direction accuracy over fixed trajectory prefixes. At $L=5$, session MAEs match those reported by~\citet{wu2026emotrack}.}
  \label{tab:prefix-evaluation}
  \begin{tabular}{lcccc}
    \toprule
    \textbf{Method} & $L=2$ & $L=3$ & $L=4$ & $L=5$ \\
    \midrule
    \multicolumn{5}{l}{\datasetqwen ($n=738$)} \\
    AIDA & 3.17/3.29/0.774 & 3.24/3.24/0.774 & 3.22/3.22/0.764 & 3.22/3.22/0.762 \\
    AIDA + history & 3.21/3.00/0.666 & 3.24/2.91/0.647 & 3.25/2.87/0.649 & 3.25/2.83/0.643 \\
    LMIQ & 2.95/3.17/0.812 & 3.01/3.16/0.795 & 2.97/3.12/0.790 & 2.98/3.08/0.779 \\
    LMIQ + history & 2.98/2.84/0.734 & 2.99/2.78/0.722 & 2.95/2.73/0.706 & 2.96/2.71/0.696 \\
    Milintsevich et al. & 2.68/3.28/0.857 & 2.71/3.38/0.837 & 2.68/3.30/0.838 & 2.70/3.25/0.828 \\
    EnsemBERT & 2.87/3.36/0.755 & 2.92/3.53/0.726 & 2.87/3.48/0.725 & 2.91/3.47/0.721 \\
    EnsemBERT (Qwen3-Emb) & 2.78/3.18/0.774 & 2.79/3.21/0.759 & 2.75/3.16/0.759 & 2.75/3.10/0.753 \\
    Lau et al. & 2.61/3.09/0.851 & 2.66/3.15/0.832 & 2.61/3.10/0.836 & 2.62/3.08/0.822 \\
    Lau et al. (Qwen3-Emb) & 2.59/3.08/0.862 & 2.63/3.12/0.835 & 2.58/3.08/0.844 & 2.59/3.06/0.833 \\
    \midrule
    \multicolumn{5}{l}{\datasetluna ($n=738$)} \\
    AIDA & 3.38/2.78/0.816 & 3.44/2.87/0.807 & 3.46/2.94/0.773 & 3.49/2.97/0.760 \\
    AIDA + history & 3.26/2.73/0.643 & 3.39/2.74/0.653 & 3.46/2.67/0.636 & 3.52/2.62/0.651 \\
    LMIQ & 3.19/2.86/0.782 & 3.26/2.94/0.766 & 3.27/2.93/0.760 & 3.28/2.93/0.759 \\
    LMIQ + history & 3.12/2.75/0.699 & 3.18/2.79/0.663 & 3.19/2.76/0.652 & 3.23/2.72/0.658 \\
    Milintsevich et al. & 2.42/2.88/0.880 & 2.48/2.95/0.868 & 2.55/3.02/0.855 & 2.56/3.01/0.849 \\
    EnsemBERT & 2.97/3.24/0.728 & 3.00/3.25/0.738 & 3.01/3.25/0.719 & 3.03/3.23/0.719 \\
    EnsemBERT (Qwen3-Emb) & 3.00/2.79/0.755 & 3.00/2.82/0.737 & 3.03/2.86/0.721 & 3.06/2.85/0.717 \\
    Lau et al. & 2.42/2.66/0.901 & 2.50/2.68/0.890 & 2.55/2.76/0.863 & 2.58/2.77/0.860 \\
    Lau et al. (Qwen3-Emb) & 2.45/2.73/0.893 & 2.53/2.75/0.886 & 2.59/2.81/0.863 & 2.60/2.82/0.856 \\
    \bottomrule
  \end{tabular}
\end{table*}

\paragraph{Transition and symptom metrics.}
Table~\ref{tab:transition-symptom-luna} reports the results of Sections~\ref{subsec:transitions} and~\ref{subsec:symptom-change} for both datasets with seed standard deviations. Symptom-level direction counts a symptom change of at least one point as correct when the predicted symptom score changes with the same sign.

\begin{table*}[t]
  \centering
  \small
  \caption{Transition- and symptom-level results as five-seed mean $\pm$ sample standard deviation. \datasetqwen contains 229 worsening and 268 improving large changes and 236 upward and 247 downward crossings of the screening cutoff; \datasetluna contains 233, 256, 205, and 223.}
  \label{tab:transition-symptom-luna}
  \begin{tabular}{lcccccc}
    \toprule
    & \multicolumn{2}{c}{\makecell{\textbf{Direction}\\\textbf{accuracy $\uparrow$}}} & \multicolumn{2}{c}{\makecell{\textbf{Threshold}\\\textbf{detection $\uparrow$}}} & \multicolumn{2}{c}{\makecell{\textbf{Symptom-level}\\\textbf{change}}} \\
    \cmidrule(lr){2-3}\cmidrule(lr){4-5}\cmidrule(l){6-7}
    \textbf{Method} & Worsening & Improvement & Upward & Downward & MAE $\downarrow$ & Direction $\uparrow$ \\
    \midrule
    \multicolumn{7}{l}{\datasetqwen ($n=738$)} \\
    AIDA & \makecell{0.746\\$\pm$0.020} & \makecell{0.776\\$\pm$0.009} & \makecell{0.205\\$\pm$0.010} & \makecell{0.249\\$\pm$0.016} & -- & -- \\
    AIDA + history & \makecell{0.708\\$\pm$0.006} & \makecell{0.587\\$\pm$0.014} & \makecell{0.136\\$\pm$0.004} & \makecell{0.087\\$\pm$0.002} & -- & -- \\
    LMIQ & \makecell{0.768\\$\pm$0.009} & \makecell{0.788\\$\pm$0.014} & \makecell{0.222\\$\pm$0.019} & \makecell{0.254\\$\pm$0.015} & -- & -- \\
    LMIQ + history & \makecell{0.727\\$\pm$0.017} & \makecell{0.670\\$\pm$0.036} & \makecell{0.113\\$\pm$0.020} & \makecell{0.104\\$\pm$0.011} & -- & -- \\
    Milintsevich et al. & \makecell{0.835\\$\pm$0.016} & \makecell{0.822\\$\pm$0.006} & \makecell{0.281\\$\pm$0.021} & \makecell{0.319\\$\pm$0.013} & \makecell{0.483\\$\pm$0.006} & \makecell{0.733\\$\pm$0.007} \\
    EnsemBERT & \makecell{0.721\\$\pm$0.019} & \makecell{0.720\\$\pm$0.013} & \makecell{0.287\\$\pm$0.025} & \makecell{0.321\\$\pm$0.028} & \makecell{0.506\\$\pm$0.021} & \makecell{0.389\\$\pm$0.014} \\
    \makecell[l]{EnsemBERT\\(Qwen3-Emb)} & \makecell{0.742\\$\pm$0.010} & \makecell{0.761\\$\pm$0.010} & \makecell{0.268\\$\pm$0.030} & \makecell{0.313\\$\pm$0.029} & \makecell{0.462\\$\pm$0.009} & \makecell{0.372\\$\pm$0.011} \\
    Lau et al. & \makecell{0.836\\$\pm$0.013} & \makecell{0.810\\$\pm$0.017} & \makecell{0.298\\$\pm$0.037} & \makecell{0.313\\$\pm$0.026} & -- & -- \\
    \makecell[l]{Lau et al.\\(Qwen3-Emb)} & \makecell{0.836\\$\pm$0.014} & \makecell{0.831\\$\pm$0.007} & \makecell{0.297\\$\pm$0.016} & \makecell{0.330\\$\pm$0.022} & -- & -- \\
    \midrule
    \multicolumn{7}{l}{\datasetluna ($n=738$)} \\
    AIDA & \makecell{0.749\\$\pm$0.013} & \makecell{0.770\\$\pm$0.008} & \makecell{0.195\\$\pm$0.010} & \makecell{0.236\\$\pm$0.012} & -- & -- \\
    AIDA + history & \makecell{0.804\\$\pm$0.013} & \makecell{0.512\\$\pm$0.008} & \makecell{0.105\\$\pm$0.006} & \makecell{0.097\\$\pm$0.004} & -- & -- \\
    LMIQ & \makecell{0.755\\$\pm$0.013} & \makecell{0.762\\$\pm$0.013} & \makecell{0.216\\$\pm$0.014} & \makecell{0.267\\$\pm$0.021} & -- & -- \\
    LMIQ + history & \makecell{0.683\\$\pm$0.024} & \makecell{0.636\\$\pm$0.005} & \makecell{0.100\\$\pm$0.014} & \makecell{0.120\\$\pm$0.008} & -- & -- \\
    Milintsevich et al. & \makecell{0.842\\$\pm$0.010} & \makecell{0.856\\$\pm$0.011} & \makecell{0.388\\$\pm$0.020} & \makecell{0.396\\$\pm$0.026} & \makecell{0.452\\$\pm$0.006} & \makecell{0.744\\$\pm$0.004} \\
    EnsemBERT & \makecell{0.731\\$\pm$0.014} & \makecell{0.707\\$\pm$0.017} & \makecell{0.294\\$\pm$0.045} & \makecell{0.352\\$\pm$0.024} & \makecell{0.469\\$\pm$0.020} & \makecell{0.350\\$\pm$0.012} \\
    \makecell[l]{EnsemBERT\\(Qwen3-Emb)} & \makecell{0.710\\$\pm$0.106} & \makecell{0.723\\$\pm$0.109} & \makecell{0.286\\$\pm$0.167} & \makecell{0.305\\$\pm$0.175} & \makecell{0.421\\$\pm$0.036} & \makecell{0.326\\$\pm$0.105} \\
    Lau et al. & \makecell{0.880\\$\pm$0.014} & \makecell{0.842\\$\pm$0.008} & \makecell{0.371\\$\pm$0.023} & \makecell{0.388\\$\pm$0.021} & -- & -- \\
    \makecell[l]{Lau et al.\\(Qwen3-Emb)} & \makecell{0.871\\$\pm$0.011} & \makecell{0.842\\$\pm$0.016} & \makecell{0.374\\$\pm$0.018} & \makecell{0.398\\$\pm$0.026} & -- & -- \\
    \bottomrule
  \end{tabular}
\end{table*}

\paragraph{Results on unseen client profiles.}
On the profile-disjoint \datasetluna split (Table~\ref{tab:luna-profile-disjoint}), the ranking difference of Section~\ref{subsec:longitudinal-results} persists: LMIQ with history has the lowest transcript-based change MAE ($2.582$) but direction accuracy $0.667$, whereas Lau et al.\ reaches $0.827$ with change MAE $3.069$. The pattern therefore does not depend on profile overlap between splits.

\paragraph{Computing infrastructure.}
Local experiments ran on a server with eight NVIDIA A100-SXM4 GPUs (80\,GB each), two Intel Xeon Gold 6346 CPUs (3.10\,GHz, 32 physical cores in total), and about 1\,TiB of RAM, running Ubuntu 22.04.5 LTS. The software environment comprised Python 3.12.13, PyTorch 2.10.0 (CUDA 12.8), cuDNN 9.10.2, Transformers 5.5.1, and vLLM 0.19.0. Qwen3.5-35B-A3B has 35B parameters, of which about 3B are active per token, and Qwen3-Embedding-8B has 8B parameters; the OpenAI models were accessed through the API, and their parameter counts are not public.

\section{Ethical Considerations}
\label{sec:ethics}

\paragraph{Intended use.}
\longcounsel is intended for research on depression assessment and longitudinal tracking. It is not a clinical tool: its clients are simulated, and a method evaluated on it should not be used to screen, monitor, or make treatment decisions for real people without clinical validation. Automated tracking of a person's mental state also requires that person's consent and clinical oversight.

\paragraph{Data and privacy.}
All released dialogues and client profiles are generated by LLMs; no real client takes part, and no real counseling conversation is released. Client profiles are derived from published counseling case reports collected in PsychEval~\citep{pan2026psycheval}, and depression scores come from public research data without personal identifiers~\citep{Makhmutova2021PredictingCI,cdc_nhanes_dpq_l_2024}. RealCBT transcripts~\citep{Wang2025FeelTD} were used, while they were publicly available, only to calibrate dialogue form and for the comparisons in Section~\ref{sec:analysis}; we release no RealCBT text. We did not run an automated audit for offensive content or personal identifiers. Because the dialogues discuss depressive symptoms by design, they can contain distressing content.

\paragraph{Licenses.}
\longcounsel is released at \datasetlink{} under the Creative Commons Attribution-NonCommercial 4.0 International (CC BY-NC 4.0) license, which preserves the noncommercial condition of PSYCHE-D and PsychEval. Appendix~\ref{apdx:asset-licenses} lists the license or access terms of every asset we use or release.

\section{Asset Licenses}
\label{apdx:asset-licenses}

Table~\ref{tab:asset-licenses} records the license or access terms of each asset we use or release. The \longcounsel release contains all three datasets, both labels, the split manifests, license files, citation instructions, and documentation of intended use and limitations. Third-party source material remains under its original terms and is not redistributed as a raw dataset.

\begin{table*}[ht]
\centering
\small
\caption{Asset licenses and access terms.}
\label{tab:asset-licenses}
\begin{tabularx}{\linewidth}{>{\raggedright\arraybackslash\hsize=.75\hsize}X
  >{\raggedright\arraybackslash\hsize=.9\hsize}X
  >{\raggedright\arraybackslash\hsize=1.35\hsize}X}
\toprule
\textbf{Asset} & \textbf{Use} & \textbf{License/access terms} \\
\midrule
\textbf{\longcounsel} & Synthetic dataset for transcript-based PHQ-8 prediction and longitudinal tracking. & Authors' original contributions: CC BY-NC 4.0. Reuse must also comply with applicable upstream terms; the intended-use statement in Appendix~\ref{sec:ethics} is guidance rather than an additional license condition. \\
\midrule
NHANES DPQ\_L~\citep{cdc_nhanes_dpq_l_2024} & Symptom co-occurrence source for \longcounsel construction. & CDC/NCHS public-use data under the NCHS Data User Agreement; no re-identification or identifiable-data linkage. \url{https://www.cdc.gov/nchs/policy/data-user-agreement.html} \\
\midrule
PSYCHE-D~\citep{Makhmutova2021PredictingCI} & Aggregate depression-change trajectory source. & CC BY-NC 4.0. Used for trajectory reconstruction; raw source files are not redistributed. \url{https://zenodo.org/records/5085146} \\
\midrule
PsychEval~\citep{pan2026psycheval} & Case/background and session-stage structures. & CC BY-NC 4.0. The source is cited and is not redistributed as a separate raw asset. \url{https://github.com/ECNU-ICALK/PsychEval/blob/main/LICENSE} \\
\midrule
RealCBT~\citep{Wang2025FeelTD} & Dialogue-form calibration and similarity comparison. & The public repository does not state a license and, as of August 4, 2026, no longer distributes raw transcripts. We cite the source and do not redistribute raw sessions. \url{https://gitlab.com/xiaoyi.wang/realcbt-dataset} \\
\midrule
Qwen models~\citep{qwen3embedding8b,qwen35_35ba3b} & Qwen3.5-35B-A3B for dialogue generation and feature extraction; Qwen3-Embedding-8B for embeddings. & Apache License 2.0 as listed on the model cards. \url{https://huggingface.co/Qwen/Qwen3-Embedding-8B}; \url{https://huggingface.co/Qwen/Qwen3.5-35B-A3B} \\
\midrule
OpenAI models & GPT-5.4-mini and GPT-5.6 Luna for dialogue generation; GPT-5.4 for snippet judging. & Accessed through the OpenAI API under the OpenAI Services Agreement and Usage Policies; no model weights are redistributed. \url{https://openai.com/policies/services-agreement/}; \url{https://openai.com/policies/usage-policies/} \\
\midrule
Evaluated methods~\citep{lee2026interpretable,rosenman2024llm,lau2023automatic,milintsevich2023towards,ravenda2025transforming} & Reimplemented or adapted comparisons. & Original papers and repositories are cited; third-party source code is not redistributed. \\
\bottomrule
\end{tabularx}
\end{table*}

\end{document}